\documentclass[10pt,twocolumn,letterpaper]{article}

\usepackage{cvpr}              % To produce the CAMERA-READY version

\usepackage{graphicx}
\usepackage{caption}
\usepackage{algorithm}
\usepackage{algorithmic}

\usepackage{amsfonts}
\usepackage{array}
\usepackage{textcomp}
\usepackage{verbatim}
\usepackage{cite}
\usepackage{booktabs}
\usepackage{bm}
\usepackage{dsfont}
\usepackage{multirow}
\usepackage[dvipsnames]{xcolor}
\usepackage[table]{xcolor}

\usepackage{titletoc}
\usepackage[framemethod=tikz]{mdframed}

\usepackage{amsmath}
\DeclareMathOperator*{\argmax}{arg\,max}

\usepackage{xspace}
\def\eg{\emph{e.g}.\xspace}
\def\ie{\emph{i.e}.\xspace}

\usepackage{newfloat}
\usepackage{listings}

\definecolor{cvprblue}{rgb}{0.21,0.49,0.74}
\usepackage[pagebackref,breaklinks,colorlinks,allcolors=cvprblue]{hyperref}

\def\paperID{*****} % *** Enter the Paper ID here
\def\confName{CVPR}
\def\confYear{2026}

\title{Robust Alignment: Harmonizing Clean Accuracy and Adversarial Robustness \\ in Adversarial Training}

\author{Yanyun Wang$^{1,2}$ \hspace{1em} Qingqing Ye$^1$ \hspace{1em} Li Liu$^2$ \hspace{1em} Zi Liang$^1$ \hspace{1em} Haibo Hu$^1$\thanks{Corresponding Author: {\tt\small haibo.hu@polyu.edu.hk}} \vspace{0.5em} \\
 $^1$HK PolyU \hspace{1em} $^2$HKUST (GZ)
}

\begin{document}
\maketitle
\begin{abstract}
Adversarial Training (AT) is one of the most effective methods for developing robust deep neural networks (DNNs). However, AT faces a trade-off problem between clean accuracy and adversarial robustness. In this work, we reveal a surprising phenomenon for the first time: Varying input perturbation intensities for training samples near decision boundaries in AT have minimal impact on model robustness. This finding directly exposes the inconsistency between accuracy and robustness score fluctuations, leading us to identify the \textbf{misalignment between input and latent spaces} as a critical driver of the robustness-accuracy trade-off. To mitigate this misalignment for harmonizing accuracy and robustness, we define \textbf{\textit{Robust Alignment}} as a new AT target, encouraging the model perception to change with input perturbations provided the final label prediction remains unchanged, which can be achieved via two novel ideas. First, we suggest a reduced and fixed perturbation intensity for those boundary samples, which facilitates the model to utilize the perturbations as learnable patterns, instead of noises that complicate decision boundaries meaninglessly. Second, we propose a \textit{\textbf{D}omain \textbf{I}nterpolation \textbf{C}onsistency \textbf{A}dversarial \textbf{R}egularization} (\textbf{DICAR}), based on rigorous theoretical derivations, which explicitly introduces semantic alignment between input and latent spaces into AT. Based on these two ideas, we end up with a new \textit{\textbf{R}obust \textbf{A}lignment \textbf{A}dversarial \textbf{T}raining} (\textbf{RAAT}) method, effectively harmonizing accuracy and robustness. Extensive experiments on CIFAR-10, CIFAR-100, and Tiny-ImageNet with ResNet-18, PreActResNet-18, and WideResNet-28-10 demonstrate the effectiveness of RAAT in improving the trade-off beyond four common baselines and a total of 14 related state-of-the-art (SOTA) works.
\end{abstract}    
\section{Introduction}

While deep neural networks (DNNs) have demonstrated impressive performance across various real-world applications, they are confirmed vulnerable to adversarial attacks~\citep{biggio2013evasion,szegedy2014intriguing}. Specifically, imperceptible perturbations added to inputs can significantly alter outputs~\citep{goodfellow2015explaining,huang2020survey}, raising serious concerns within the public and community~\citep{wang2020improving}. Several defense techniques, such as Defense Distillation~\citep{papernot2017practical}, Feature Squeezing~\citep{xu2017feature}, Randomization~\citep{xie2018mitigating} and Input Denoising~\citep{guo2018countering,liao2018defense}, have been proposed to enhance adversarial robustness of DNNs, but most of them have subsequently proven ineffective against advanced adaptive attacks~\citep{athalye2018obfuscated,tramer2020adaptive}. Currently, Adversarial Training (AT)~\citep{goodfellow2015explaining,madry2018towards} is recognized as one of the most effective methods to train inherently robust DNNs~\citep{athalye2018obfuscated,dong2020benchmarking}, with several specific algorithms like PGD-AT~\citep{madry2018towards}, TRADES~\citep{zhang2019theoretically} and MART~\citep{wang2020improving} serving as solid benchmarks.

However, recent studies have identified a \textbf{trade-off} in AT between adversarial robustness and clean accuracy~\citep{tsipras2019robustness,zhang2019theoretically,raghunathan2019adversarial,raghunathan2020understanding,wang2020once,bai2021recent,yin2023push,liu2025parameter}. Specifically, improvements in robustness through AT often come at the expense of reduced model accuracy compared to standard training. This trade-off can significantly degrade the user experience in benign cases and limit the adoption of AT in real-world applications. A widely recognized factor contributing to this issue is that, to uniformly achieve prediction consistency under adversarial perturbations within the $\epsilon$-ball of each natural data point, AT tends to learn more complex decision boundaries than standard training~\citep{dong2022exploring,rade2022reducing,cheng2022cat,yang2022one,yin2023push,liu2025parameter}, which harms the generalization ability of the model on unseen data. To mitigate this, previous works have explored strategies for learning those \textbf{boundary samples} near decision boundaries more appropriately, but their opinions remain divided. GAIRAT~\citep{zhang2021geometry} and MAIL~\citep{liu2021probabilistic} believe boundary samples are more critical and should be learned with larger weights or enhanced patterns, while MMA~\citep{ding2020mma}, HAT~\citep{rade2022reducing} and TE~\citep{dong2022exploring} hold that reducing perturbations or weights for boundary samples would benefit AT generalization. 
% Thus, the optimal strategy for learning boundary samples remains an open question in AT.

Given the current mixed views on the learning of boundary samples in AT, we first restudy the impact of boundary samples to model accuracy and robustness. Surprisingly, we reveal that for those boundary samples correctly classified yet close to the decision boundary, adopting different perturbation intensities for them in AT can \textbf{hardly} impact the final robustness learned by the model, which implies that certain perturbation information within the input is not effectively learned by the model and reflected to the latent representation as expected. Instead, it only introduces noise that meaninglessly complicates decision boundaries, degrading the clean accuracy and resulting in semantic \textbf{misalignment} between input and latent representation.

\begin{figure}[htbp]
    \centering
    \includegraphics[width=\linewidth]{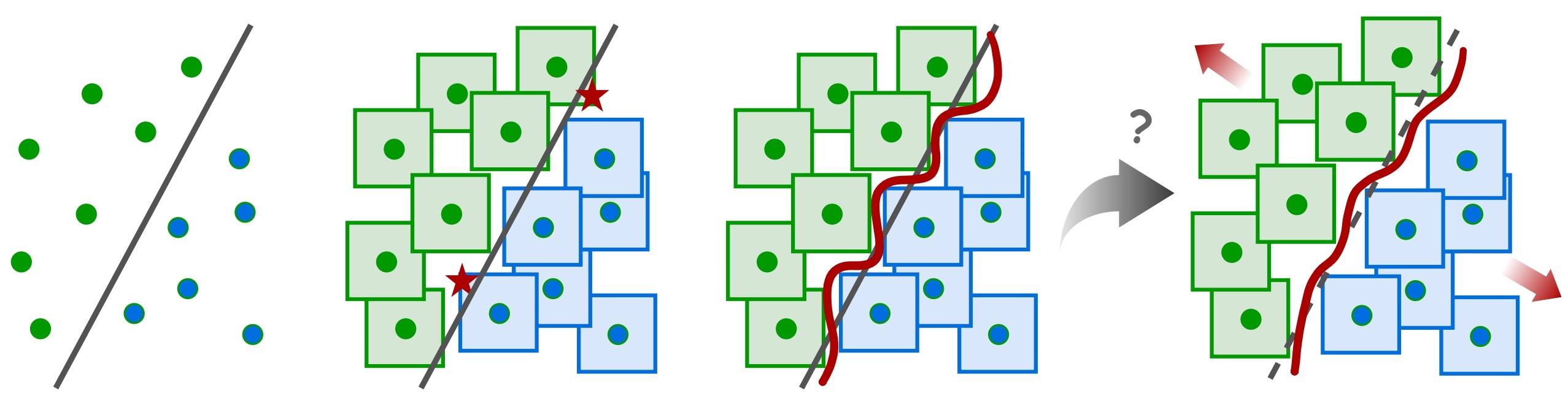}
    \caption{A high-level glance of motivation. \textbf{Left:} An illustration for the effectiveness of AT in improving model robustness (taken from \citet{madry2018towards}). This implies that AT can naturally result in a significantly more complicated decision boundary than clean training, which introduces a certain risk of semantic misalignment between input and latent representation, especially for boundary samples. \textbf{Right:} An intuitive target proposed. In turn, facilitating the alignment of input and latent spaces is expected to smooth the decision boundary and benefit the dynamics of model learning in pushing different class clusters away from each other.}
    \label{fig:AT_generalization}
\end{figure}

In this work, we identify such \textbf{misalignment} as a key reason for the current trade-off problem. Figure~\ref{fig:AT_generalization} demonstrates the intuition behind this view. Accordingly, we define a new target for AT named \textbf{\textit{Robust Alignment}}, aiming at aligning the input and latent space for harmonizing accuracy and robustness. Specifically, different from the current knowledge viewing adversarial perturbations as noises, \textit{Robust Alignment} encourages the change of model perception along with perturbations added into the input as long as the final label prediction remains unchanged. Our realization of \textit{Robust Alignment} consists of two new ideas. First, we reduce the perturbation intensity of boundary samples to a fixed ratio in AT, such that certain information within adversarial perturbations can more easily form learnable patterns for the model, instead of just becoming training noise. Second, we propose a novel \textit{\textbf{D}omain  \textbf{I}nterpolation  \textbf{C}onsistency  \textbf{A}dversarial  \textbf{R}egularization} (\textbf{DICAR}), explicitly introducing semantic alignment between input and latent spaces into AT. It encourages the prediction of interpolation of the adversarial samples respectively from two augmentations to be consistent with the interpolation of predictions of the two augmentations. The mechanism can be intuitively understood as utilizing aligned domains beyond just single points or pairs of adversarial instances as in existing works. Rigorous theory proves that DICAR acts as a regularizer on the derivatives of all orders to reduce the misalignment. 

Based on these two ideas, we propose \textit{\textbf{R}obust \textbf{A}lignment \textbf{A}dversarial \textbf{T}raining} (\textbf{RAAT}), involving a novel general risk as our new AT objective and two specific surrogate methods to optimize it. Extensive experiments on CIFAR-10, CIFAR-100, and Tiny-ImageNet datasets with ResNet-18, PreActResNet-18, and WideResNet-28-10 architectures demonstrate the effectiveness of RAAT in concurrently improving clean accuracy and robustness compared with four common AT benchmarks and a total of 14 advanced related works and state-of-the-art (SOTA) methods on the accuracy-robustness trade-off problem. 
% The code is available in our GitHub repository\footnote{During the review process, we temporarily hold off on releasing the code to the public and instead provide it as Supplementary Material separately. The code implementation is based on \citet{tack2022consistency}.}.
% PGD-AT, TRADES, MART, and Cons-AT.

\begin{figure*}[htb]
  \centering
  \subfloat[\small Robust acc - boundary sample]
  {\includegraphics[width=0.337\textwidth]{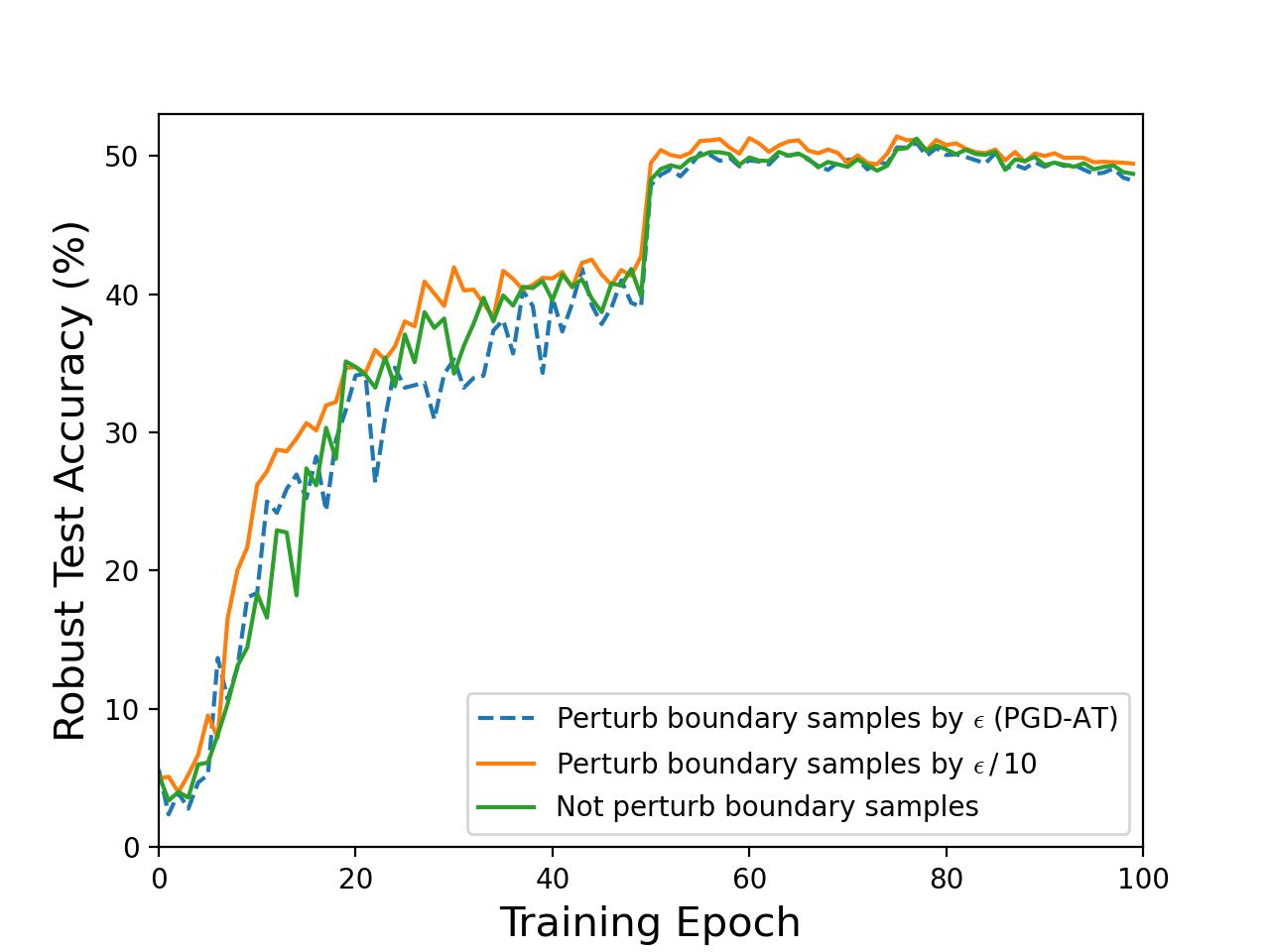}}
  \subfloat[\small Robust acc - non-boundary sample]
  {\includegraphics[width=0.337\textwidth]{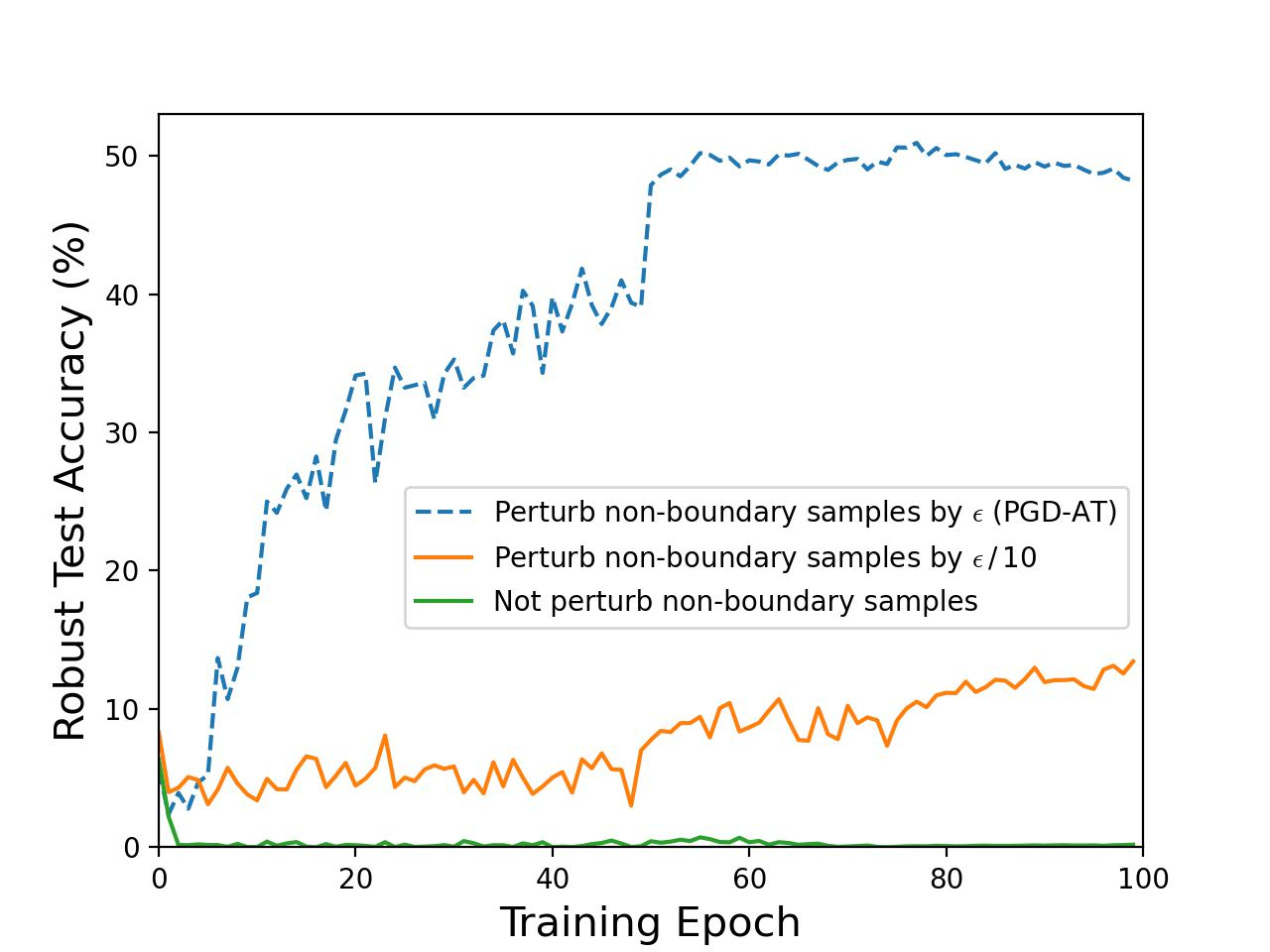}}
  \subfloat[\small Robust acc - partition strategy]
  {\includegraphics[width=0.337\textwidth]{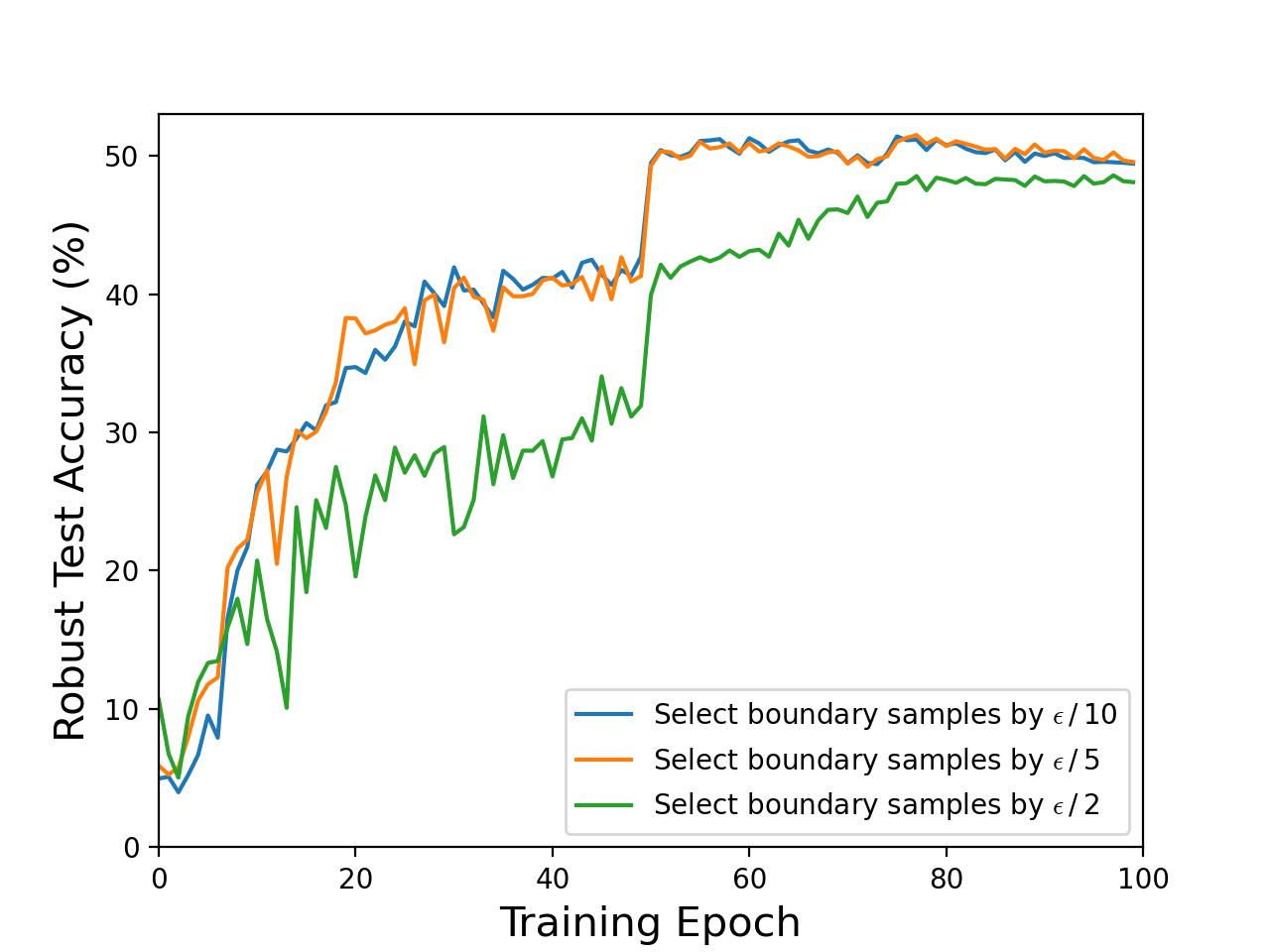}} \\
  \subfloat[\small Clean acc - boundary sample]
  {\includegraphics[width=0.337\textwidth]{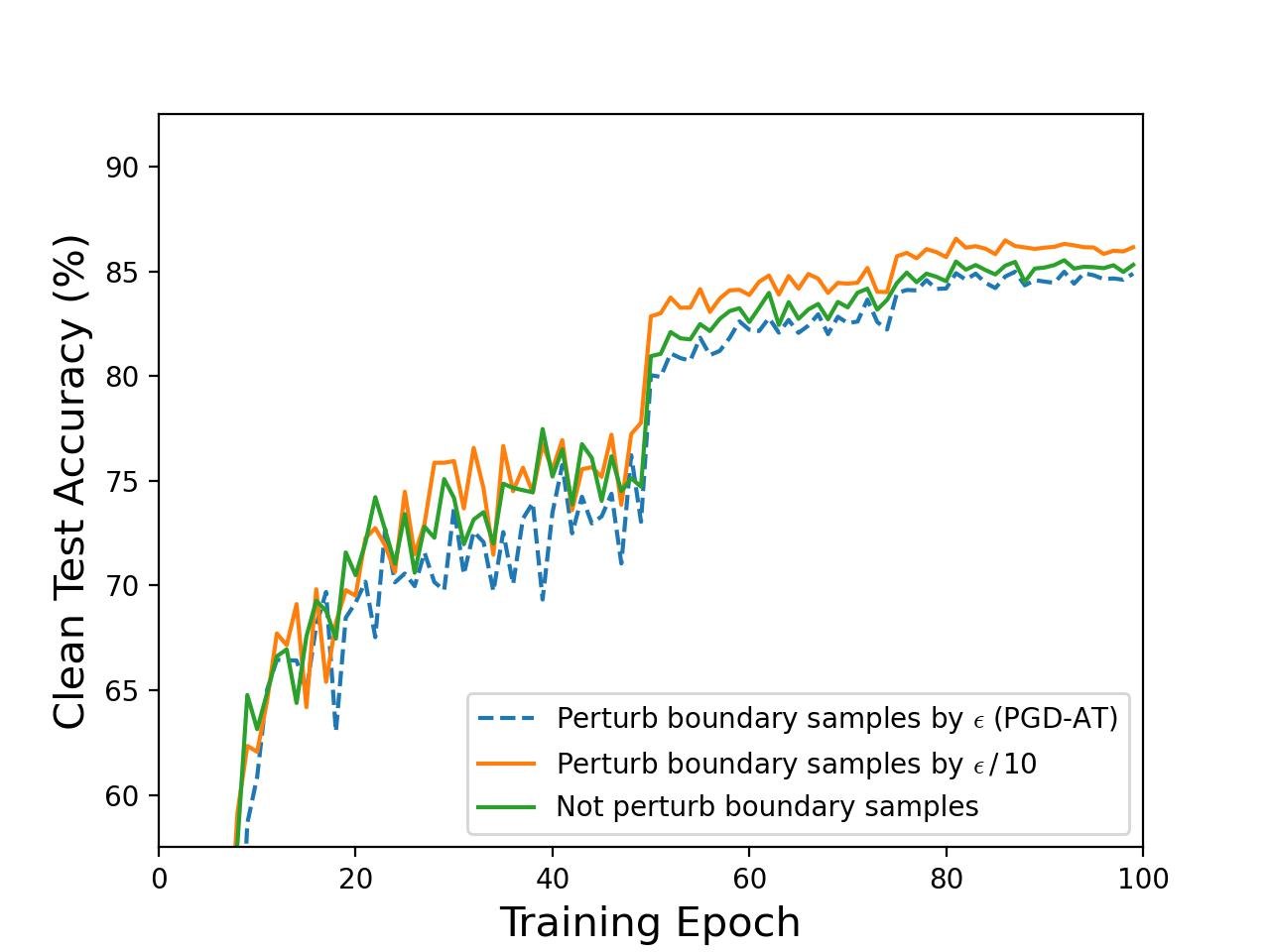}}
  \subfloat[\small Clean acc - non-boundary sample]
  {\includegraphics[width=0.337\textwidth]{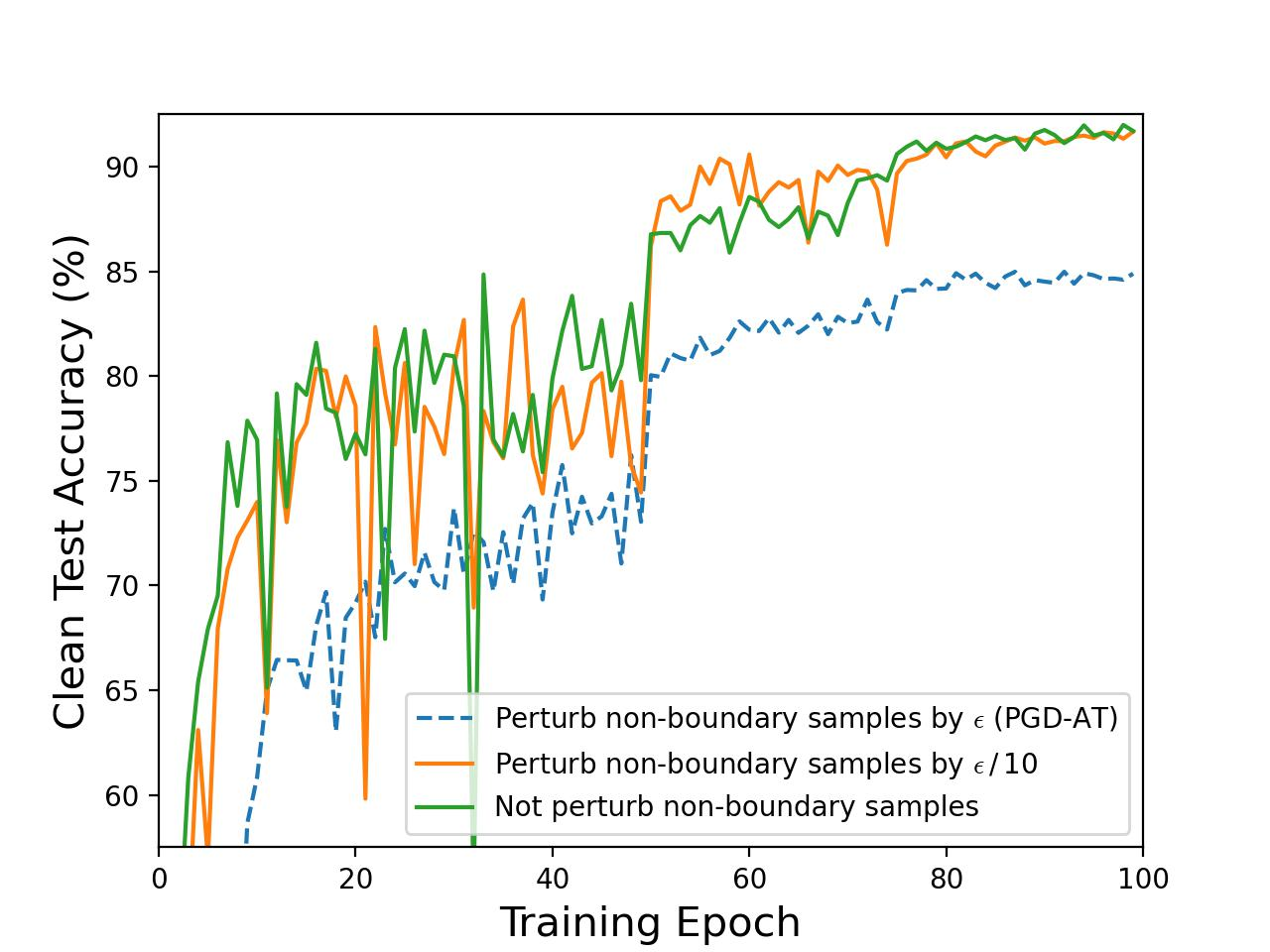}}
  \subfloat[\small Clean acc - partition strategy]
  {\includegraphics[width=0.337\textwidth]{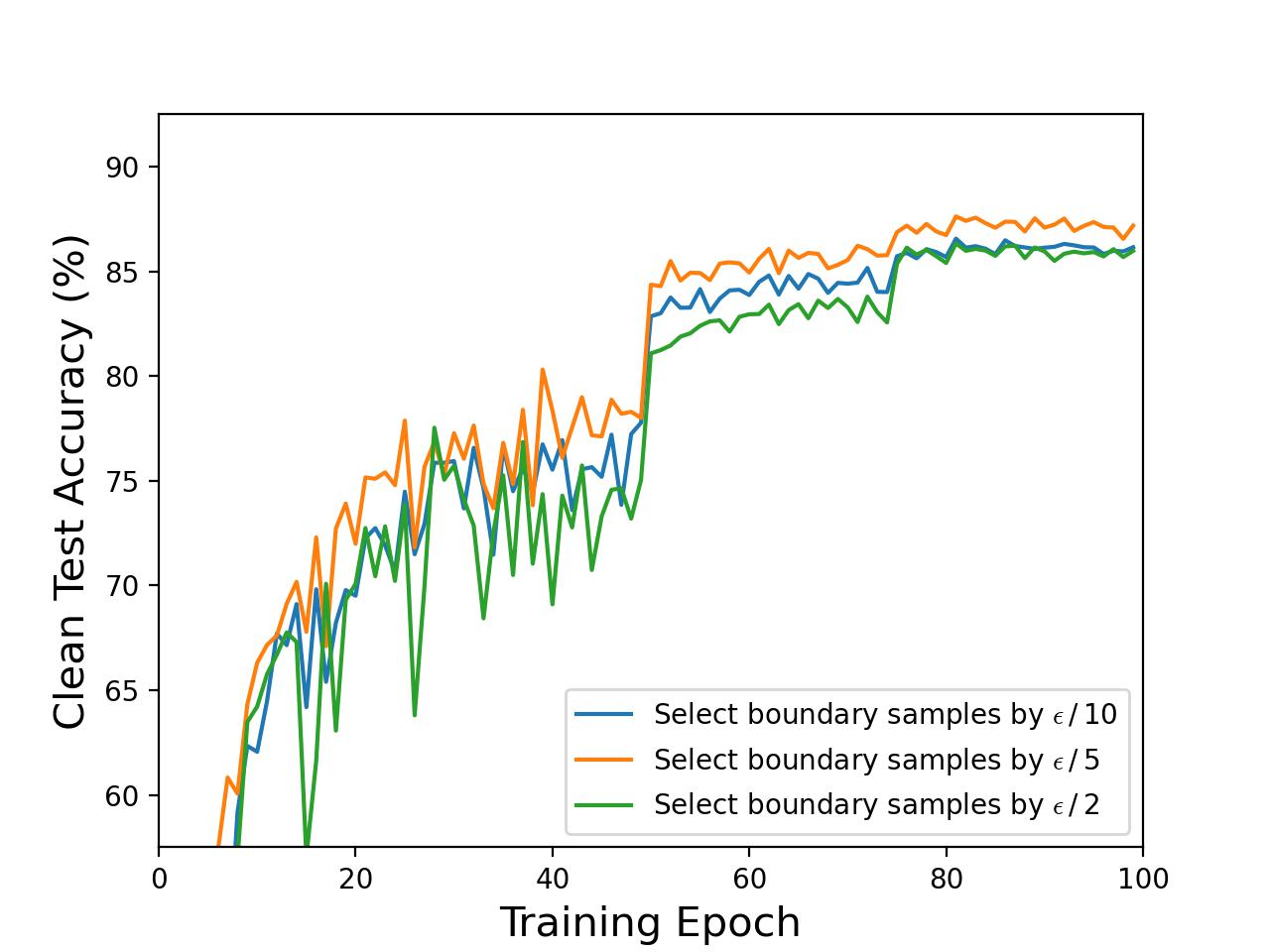}}
  \vspace{0.2em}
  \caption{The figures show a series of proof-of-concept experiments on the role of boundary and non-boundary samples in AT. \textbf{(a) and (d):} Reducing the intensity of perturbations applied to the boundary samples has basically \textbf{no influence} on the model robustness, while appropriate perturbation \textbf{does help} to improve the clean accuracy. \textbf{(b) and (e):} In contrast, fully perturbing the non-boundary samples within the $\epsilon$-ball is \textbf{vital} to learn robustness. \textbf{(c) and (f):} The partition of the boundary and non-boundary samples matters and an appropriate strategy can benefit both robustness and clean accuracy.}
  \label{fig:motivation_boundary_samples}
\end{figure*}

\section{Robust Alignment: A New Target of AT}\label{sec:robust_alignment}

In this section, we aim to answer a question: \textbf{Can we directly harmonize accuracy and robustness in AT?} Beginning with further study of boundary samples, we first identify an underlying cause of the accuracy-robustness trade-off problem: The \textbf{misalignment between input features and latent representations} of the model. Then accordingly, we propose a new target of AT named \textbf{\textit{Robust Alignment}} to mitigate such an undesirable side-effect it causes.

\subsection{Restudying Boundary Samples in AT}\label{subsec:boundary_sample}

The current accuracy-robustness trade-off is widely attributed to the more complicated decision boundary caused by AT~\citep{dong2022exploring,rade2022reducing,cheng2022cat,yang2022one}, with many mainstream solutions focusing on the ``boundary samples'' that are correctly classified but near the decision boundary~\citep{zhang2021geometry,liu2021probabilistic,ding2020mma,rade2022reducing,zhang2020attacks,dong2022exploring}. More details can be found in Appendix~\ref{subapp:backg_tradeoff}. The principle is quite natural because, compared with those embedded close to the center of class clusters, boundary samples are expected to have a greater impact on the establishment of decision boundary, as well as be more vulnerable when the boundary is built undesirable. However, there are opposite views on how to appropriately treat boundary samples for better AT among existing works~\citep{zhang2021geometry,liu2021probabilistic,ding2020mma,rade2022reducing,zhang2020attacks,dong2022exploring}. Thus, here we first restudy the impact of boundary samples in AT to determine the standing point of our solution.

Based on \citet{pang2021bag}, we conduct proof-of-concept experiments on CIFAR-10 with ResNet-18, as shown in Figure~\ref{fig:motivation_boundary_samples}. First, we train a baseline model (\ie, blue dashed lines) for 100 epochs using PGD-AT with default settings, except following the suggestion from MMA~\citep{ding2020mma} to include misclassified samples only in a clean manner. We then set a threshold \textit{w.r.t.} the perturbation budget $\epsilon$ (\eg, given the default $\epsilon \!=\! 8/255$, we adopt $\epsilon/10 \!=\! 0.8/255$ as the threshold) to identify the boundary samples as those natural samples that can still be successfully attacked with weakened perturbations under the threshold. To reveal and compare the contributions of boundary and non-boundary samples to AT, we respectively adjust the adversarial perturbations applied to them to either the threshold or zero (\ie, orange and green lines in Figure~\ref{fig:motivation_boundary_samples} (a), (b), (d) and (e)).

Surprisingly, we found that \textbf{varying perturbation intensities for boundary samples have little effect on the final robustness} of the model, while appropriate perturbations can improve clean accuracy. In contrast, reducing perturbation intensity for non-boundary samples significantly degrades the effectiveness of AT in achieving robustness. This result updates the existing knowledge. For the first time, we suggest from a statistics perspective that, it means \textbf{little} for the learning of robustness to manipulate the weights and perturbation intensities of certain boundary samples. More importantly, we directly reveal the \textbf{misalignment} of accuracy and robustness \textit{w.r.t.} the boundary samples in AT. That is, \textbf{different perturbations for boundary samples lead to varying accuracy but fixed robustness}, which inspires the proposal of our new target of AT called \textit{Robust Alignment}.

% Despite some previous works with adaptive $\epsilon$ like MMA~\citep{ding2020mma} utilize similar principle, our work further shows that a fixed reduction factor suffices for all the boundary samples, avoiding additional computational costs from adaptive algorithms. 

\subsection{Harmonizing Clean Accuracy and Adversarial Robustness in AT through Robust Alignment}\label{subsec:robust_alignment}

Intuitively, as shown in Figure~\ref{fig:AT_generalization}, it is not difficult to find an ideal model harmonizing accuracy and robustness. The key points can be summed up as: The decision boundary of such a model should be less complicated and farther from natural data points than existing ones, so that the generalization ability is guaranteed and the risk of altering the semantic information of samples after adversarial perturbation is limited. Then why do we suffer from the existing trade-off? More specifically, \textbf{why does conventional AT build models in the current mode (\ie, as Figure~\ref{fig:AT_generalization} (left)) instead of the more ideal one (\ie, as Figure~\ref{fig:AT_generalization} (right))?} Based on the results in Section~\ref{subsec:boundary_sample}, we believe an important cause of such a current failure is that, the conventional approach to perturb boundary samples like non-boundary ones in AT is not sufficiently effective, especially regarding the generalization ability. More specifically, \textbf{fully perturbed boundary samples may lack statistically significant and learnable patterns for robust generalization} on unseen data. So unlike boundary samples in standard training that play a role in the stable separation of class clusters, these degraded adversarial boundary samples can hardly support the general directionality of robust optimization. Instead, they just result in fragmented robust domains embedded with each other, unnecessarily complicating the decision boundary. 

\begin{figure}[htbp]
  \centering
  % \vspace{-1em}
  \subfloat[\small]
  {\includegraphics[width=0.19\textwidth]{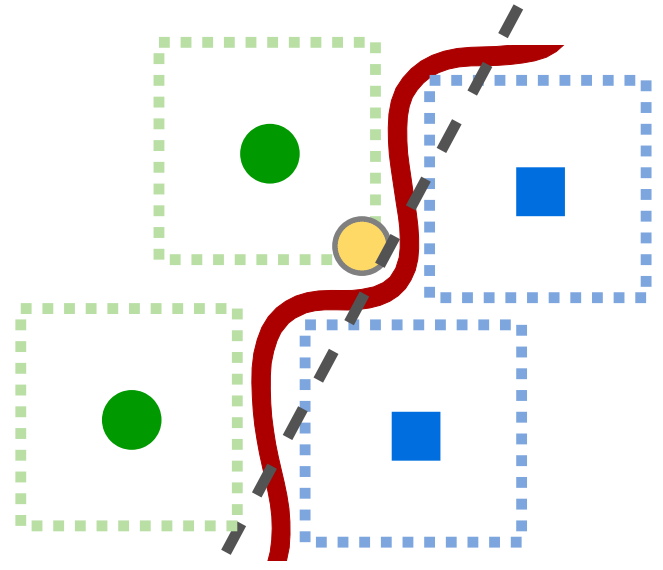}}
  \hspace{1em}
  \subfloat[\small]
  {\includegraphics[width=0.19\textwidth]{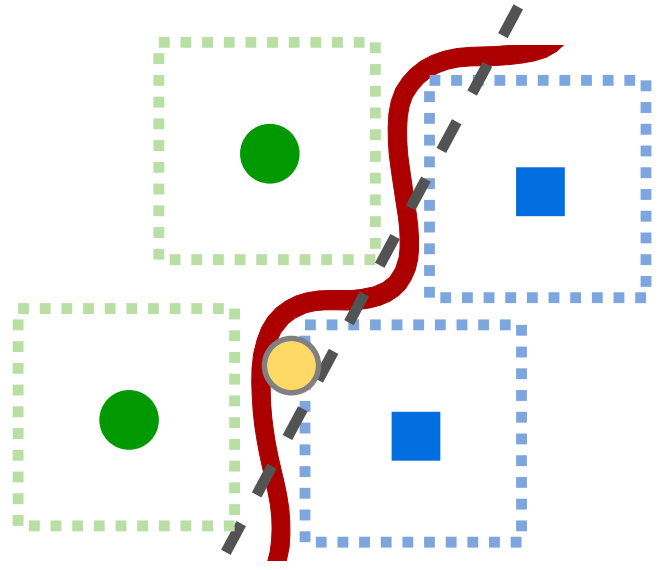}}
  \caption{An example demonstrating the generalization issue of AT and the necessity of \textit{Robust Alignment}. Red lines are the projections of decision boundary on the input space, while gray dashed lines represent the potential ground truth. \textbf{(a):} AT aims to build robust domains (dashed box) around each training data point, so that when a test sample (yellow circle, either benign or adversarial) falls into them, it can be robustly classified. \textbf{(b):} But as a side effect, this complicates the decision boundary, resulting in certain areas where input and latent manifold are \textbf{misaligned} (\ie, the fragmented areas between two boundaries). As a consequence, any input falling into these areas will be wrongly perceived as the opposite class, leaving room for unseen adversarial samples.}
  \label{fig:robust_alignment}
\end{figure}

\textbf{Why do we identify this issue as a cause of the accuracy-robustness trade-off in AT?} In essence, this can be explained from the perspective of \textbf{semantic inconsistency between input features and latent representations}. Specifically, as illustrated in Figure~\ref{fig:robust_alignment}, due to this issue, although conventional AT does improve the robustness around the training data points, for certain unseen data points (\eg, the ones out of the constructed robust domain), AT may also aggravate the misalignment between their input and latent spaces, increasing the risk of altering the semantic information of their latent representation under adversarial perturbation (while the semantic information of their input features actually remains unchanged). In other words, that means the robustness learned through conventional AT does not always generalize well as expected to unseen adversarial samples. 

% \begin{figure}[htbp]
%   \centering
%   \hspace{-1em} 
%   \begin{subfigure}{0.49\linewidth}
%     \includegraphics[width=1\linewidth]{Figures/.png}
%     \caption*{(a)}
%   \end{subfigure}
%   \begin{subfigure}{0.49\linewidth}
%     \includegraphics[width=1\linewidth]{Figures/}
%     \caption*{(b)}
%   \end{subfigure}
%   \caption{}
%   \label{fig:}
% \end{figure}
% AT在align training sample的各种adv sample的input与manifold的同时，反而增加了test set中不被包含在train sample所构建的这些鲁棒域中的那些sample的input与manifold之间的misalignment，导致它们被微小扰动变为adv sample的几率反而增大了

In response, we propose a new additional target for AT named \textbf{\textit{Robust Alignment}}, directly \textbf{facilitating the semantic alignment} between input and latent representation, to \textbf{reduce the inconsistency} between input and latent class clusters, thus \textbf{alleviating the trade-off} between accuracy and robustness. Previously, the semantic deviation between input and latent manifolds has been studied as a basis for the existence of adversarial samples. For instance, \citet{wang2022tsfool} explained that such deviation allows samples similar in input features to have quite different latent representations, increasing the risk of small input perturbations significantly altering model perception. This also supports the effectiveness of the proposed \textit{Robust Alignment}, because it directly aims at the existence of adversarial samples, fulfilling the ultimate goal of robustness. Nevertheless, to our knowledge, our work is the \textbf{first} to explicitly attribute the lack of harmonization between accuracy and robustness in AT to the semantic misalignment between input and latent spaces.

% 目标是x->x'时，predicted label不变，但是prediction (probablity)要变
\vspace{0.5em} \noindent \textbf{Definition 1} (Robust Alignment)\textbf{.} \hspace{1pt} \textit{Given a target model $\theta$ with depth $d_\theta$, let $h_\theta(x)_{\xi}$ denote the hidden representation after the $\xi$-th layer of $\theta$, for any input instance $\mathbf{x}$ and the corresponding adversarial sample $\mathbf{x}'$ with $\Delta = x^{\prime} - x$, the target of Robust Alignment can be formulated as:}
\begin{equation*}\label{eq:robust_alignment}
\begin{aligned}
    & \forall \,\, \xi \in \{ 1, ..., d_\theta \} \text{ and } \mu \in [0, 1], \\
    & \,\,\,\,\,\,\,\,\,\, h_\theta(x + \mu \!\cdot\! \Delta)_{\xi} - h_\theta(x)_{\xi} = \mu \cdot (h_\theta(x')_{\xi} - h_\theta(x)_{\xi}).
\end{aligned}
\end{equation*}
% \vspace{0.5em} \noindent \textbf{Definition 1} (\textit{\textbf{Robust Alignment}})\textbf{.} \hspace{1pt} \textit{Provided an AT task with target model $\theta$ with depth $\Xi$, let $h_\theta(x)_{\xi}$ denote the hidden representation after the $\xi$-th layer of $\theta$, for any input instance $\mathbf{x}$ and the corresponding adversarial sample $\mathbf{x}'$ with $\Delta = x^{\prime} - x$, the target of Robust Alignment can be formulated as:}
% \begin{equation*}\label{eq:robust_alignment}
% \begin{aligned}
%     & \forall \,\, \xi \in \{1, ..., \Xi\} \text{ and } \mu \in [0, 1], \\
%     & \,\,\,\,\,\,\,\,\,\, h_\theta(x + \mu \!\cdot\! \Delta)_{\xi} - h_\theta(x)_{\xi} = \mu \cdot (h_\theta(x')_{\xi} - h_\theta(x)_{\xi}).
% \end{aligned}
% \end{equation*}

\textit{Robust Alignment} is \textbf{not} an optimization objective directly available for model learning. Instead, it is a conceptual definition containing a \textbf{new philosophy} to be noted in AT. That is, although adversarial robustness requests \textbf{prediction} consistency within the perturbation budget, it does not directly mean \textbf{perception} consistency. On the contrary, as long as the final output label may not be overturned, the model should be encouraged to change its perception of different adversarial variants of the same sample, especially when the variants approach the decision boundary. In this way, the input features and latent representations can be better aligned, reducing the risk from unseen adversarial samples as shown in Figure~\ref{fig:robust_alignment}. However, conventional AT simply adopts hard-label learning, lacking explicit distinction between the learning objectives of different adversarial variants crossing training epochs, which is afraid to make the model confused about their patterns and degrade their statistical learnability. To the best of our knowledge, this is the \textbf{first} work explicitly identifying this issue as a cause of the accuracy-robust trade-off in AT.
\section{Approaching Robust Alignment via Domain Interpolation Consistency Regularization}\label{subsec:DICAT}

Compared with conventional AT adopting hard-label learning, some recently proposed AT methods based on Consistency Regularization (CR) involve prediction similarity as a softer objective~\citep{dong2022exploring,zhang2022alleviating,tack2022consistency}. According to the analysis in Section~\ref{sec:robust_alignment}, this is expected to somewhat mitigate the misalignment between input and latent spaces, which might also be an underlying reason for their effectiveness. However, as detailed in Appendix~\ref{subapp:backg_CR}, these previous works mainly focus on robust overfitting, and there are few theoretical and empirical contributions achieved \textit{w.r.t.} the accuracy-robustness trade-off problem. In this section, we first prove the \textbf{theoretical validity} of CR in solving this problem from the perspective of information theory, followed by proposing our new approach, \textit{\textbf{D}omain  \textbf{I}nterpolation  \textbf{C}onsistency  \textbf{A}dversarial  \textbf{R}egularization} (\textbf{DICAR}), which utilizes the principle of CR to establish a novel additional learning objective to \textbf{approach} the newly defined \textit{Robust Alignment}.

\begin{figure*}[htbp]
    \centering
    \includegraphics[width=0.99\linewidth]{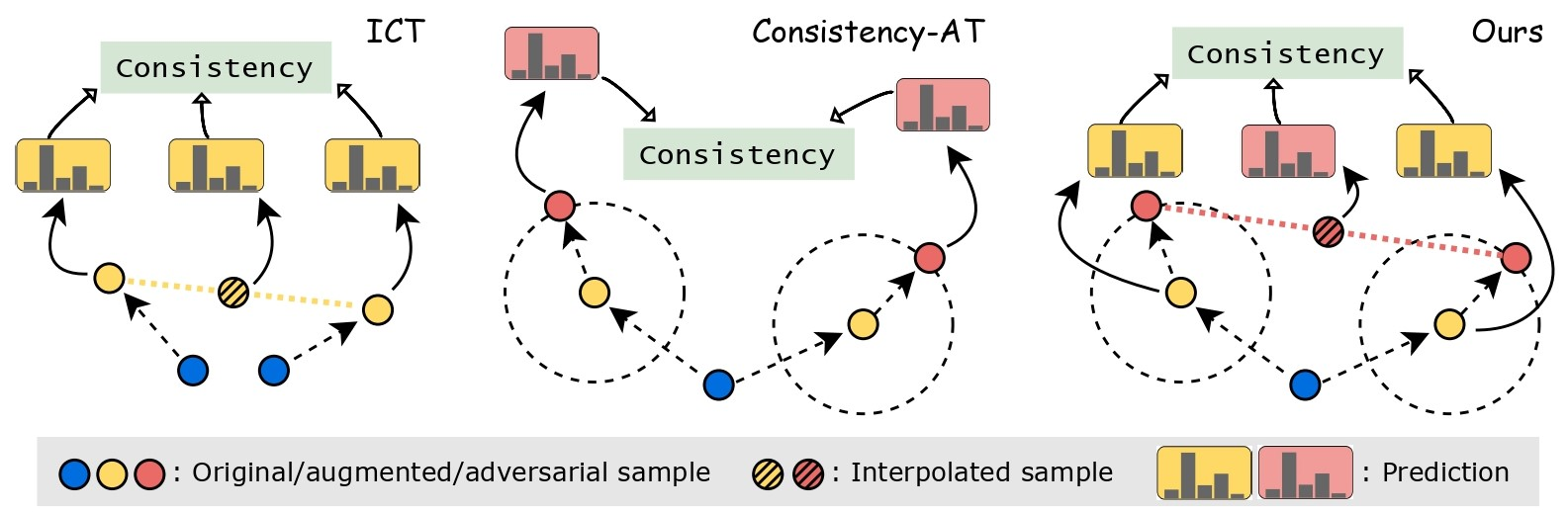}
    \caption{The figure compares the basic ideas of three different CR-based approaches. \textbf{Left:} ICT expects the model to produce a label on an interpolated sample to match the interpolation of predictions of the two corresponding original samples. \textbf{Middle:} Cons-AT expects the adversarial samples from two different augmentations of a specific instance to be consistent with each other. \textbf{Right:} Given two different augmentations from a specific original sample, our approach expects the prediction of interpolation of the two adversarial samples from them to be consistent with the interpolation of their predictions.}
    \label{fig:consistency_compare}
\end{figure*}

\subsection{Theoretical Validity of CR on Harmonizing Accuracy and Robustness in AT}

To demonstrate the theoretical potential of CR in harmonizing accuracy and robustness in AT, we examine the gap between the expected lower bounds of standard and robust generalization error to estimate the risk of disharmony. Specifically, we adopt \textit{Gaussian Model}~\citep{schmidt2018adversarially}, which reveals the information-theoretic gap between standard and robust generalization from the perspective of sample complexity.

\vspace{0.6em} \noindent \textbf{Definition 2} (Gaussian Model~\citep{schmidt2018adversarially})\textbf{.} \hspace{1pt} \textit{Let $\theta^\star \in \mathbb{R}^d$ be the per-class mean vector and let $\sigma > 0$ be the variance parameter. Then the $(\theta^*, \sigma)$-Gaussian model is defined by the following distribution over $(x,y) \in \mathbb{R}^{d} \times \{\pm1\}$: First, draw a label $y \in \{\pm1\}$ uniformly at random. Then sample the data point $x \in \mathbb{R}^{d}$ from $\mathcal{N} (y \cdot \theta^{\star}, \sigma^{2}I)$.}
\vspace{0.7em}

In \citet{schmidt2018adversarially}, this setting is used to derive an example algorithm achieving fixed and arbitrary accuracy with, for simplicity, just a single sample. Here we extend the \textit{Gaussian Model} to the context of CR as Definition 3, based on which we propose Theorem 1, stating the theoretical potential of CR in harmonizing accuracy and robustness in AT. This also lays a solid foundation for the next subsection, where we propose a novel CR-based method to approach the \textit{Robust Alignment}.

\vspace{0.6em} \noindent \textbf{Definition 3} (Gaussian CR)\textbf{.}\footnote{
Although being theoretically similar to \textit{Gaussian UAT-FT} \citep{alayrac2019labels}, \textit{Gaussian CR} is proposed with a different real-world concern. In semi-supervised learning, CR can utilize additional unlabeled data that are \textit{i.i.d.} with the labeled ones to learn better generalization. But in AT, a conventional fairness assumption is not to include any additional data~\citep{pang2021bag}. So existing works typically adopt $n = m$ with $x_{n+i}$ augmented from $x_{i}$ to approximate CR in practice~\citep{dong2022exploring,zhang2022alleviating}.
} \hspace{1pt} \textit{Given $n$ labeled samples $(x_1, y_1), \ldots, (x_n, y_n)$ and $m$ unlabeled ones $x_{n+1}, \ldots,$ $x_{n+m}$, with $\theta \!=\! \sum_{i=1}^{n} y_i x_i$ denoting the sample mean vector on the labeled samples, a Gaussian CR estimator can be defined as $\tilde{\theta} \!=\! \sum_{i=n+1}^{n+m} \tilde{y}_i x_i$, provided that $\tilde{y}_i \!=\! f_{\theta}(x_i)$.}

\vspace{0.7em} \noindent \textbf{Theorem 1.} \textit{Considering a $(\theta^*, \sigma)$-Gaussian model with $\|\theta^*\|_2 \!=\! \sqrt{d}$ and $\sigma \!\leq\! \frac{1}{32} d^{1\!/4}$, we have an expected gap between the lower bounds of standard and robust generalization error increasing with $O(\sqrt{d}/\log d)$. While provided the corresponding Gaussian CR estimator $\tilde{\theta}$, the gap can be filled at the cost of $O(\sqrt{d})$ (unlabeled) data points.}

\vspace{0.5em} \noindent \textit{Proof.} We defer the proof to Appendix~\ref{app:theorem1}.

\subsection{New CR-based Method for Robust Alignment}\label{subsec:DICAR}

To realize the philosophy of \textit{Robust Alignment} in practice, in this section, we propose a novel CR-based approach named DICAR, explicitly enhancing semantic alignment between input and latent spaces, which is expected to better mitigate the current accuracy-robustness trade-off. 

\vspace{0.5em} \noindent \textbf{Definition 4} (Domain Interpolation Consistency Adversarial Regularization (DICAR))\textbf{.} \hspace{1pt} \textit{Provided an AT task with target model $f_\theta$, loss function $\ell$ and robust budget $\epsilon$, for any input instance $x$, given $\dot{x}$ and $\ddot{x}$ as its two different augmented variants, $\dot{x}'$ and $\ddot{x}'$ as the corresponding adversarial samples, and $\hat{\lambda} = 1 - \lambda$, the DICAR term can be defined as:}
\begin{equation*}\label{eq:DICAR}
    \min_\theta \mathbb{E} \Big\{\! \max_{\! \Vert \dot{x}' \!-\! \dot{x} \Vert \leq \epsilon, \Vert \ddot{x}' \!-\! \ddot{x} \Vert \leq \epsilon \!} \! \ell \Big(\! f_\theta( \lambda \dot{x}' \!+\! \hat{\lambda} \ddot{x}' \!) \big(\! \lambda f_\theta(\dot{x}) \!+\! \hat{\lambda} f_\theta(\ddot{x}) \big) \!\Big) \!\Big\}.
\end{equation*}

Given a natural sample and its two augmentations, the proposed DICAR encourages the prediction of the interpolation of adversarial samples respectively from the two augmentations to be consistent with the interpolation of the predictions of the two augmentations. Intuitively, this is comparable in specific implementation to Interpolation Consistency Training (ICT)~\citep{verma2022interpolation} and Cons-AT~\citep{tack2022consistency}. Utilizing the idea of \textit{Mixup}~\citep{zhang2019mixup}, ICT adds interpolations of data points and expects the model prediction to be consistent with the interpolated predictions of the corresponding data points. While as detailed in Appendix~\ref{subapp:backg_CR}, Cons-AT expects the adversarial samples from two different augmentations of a specific instance to be consistent with each other. Figure~\ref{fig:consistency_compare} intuitively compares our DICAR with ICT and Cons-AT.

The effectiveness of DICAR can be demonstrated from two perspectives. On the one hand, DICAR \textbf{explicitly supports consistent domains} between interpolated samples and the corresponding two augmentations, facilitating semantic alignment of input and latent representation within the certain range of the corresponding $\epsilon$-ball to be optimized. Intuitively, such aligned ``domains'' will inject a stronger inductive bias into the target model than the individual adversarial samples (\ie, ``points'') relied on by conventional AT and the consistency relations (\ie, ``edges'') utilized by previous CR-based methods like Cons-AT. On the other hand, formally, as demonstrated in Theorem 2, the proposed DICAR term can serve as a regularizer on the derivatives of all orders. That is, DICAR \textbf{encourages any small changes within the input to be faithfully perceived and learned by the target model}. Thus, with the additional DICAR term, the target model will learn to produce similar \textbf{predictions} under different \textbf{perceptions} of various adversarial variants from the same benign sample, instead of being forced to perceive different inputs as similar latent representations, just as in conventional AT. As a result, DICAR helps to reduce the misalignment between input and latent features, effectively meeting our new target suggested by \textit{Robust Alignment}.
% preventing significant prediction changes \textit{w.r.t.} input~\citep{verma2022interpolation}. 
% as well as the risk of overfitting.

\vspace{0.5em} \noindent \textbf{Theorem 2.} \textit{Let $x \in \mathbb{R}^d$ and $f_\theta$ be real analytic, given $\Delta = x^{\prime} - x$ with $x^{\prime}$ denotes the corresponding adversarial sample of $x$, and a small perturbation budget $\epsilon \to 0$ such that $\dot{x}' \to \dot{x}$ and $\ddot{x}' \to \ddot{x}$, then for any $K \in \mathbb{N}^+$, we have:}
\begin{equation*}\label{eq:theorem2}
\begin{aligned}
    & \ell \hspace{0.1em} \Big( f_\theta(\lambda \dot{x}' + \hat{\lambda} \ddot{x}') \hspace{0.1em} \big( \lambda f_\theta(\dot{x}) + \hat{\lambda} f_\theta(\ddot{x}) \big) \Big) \\
    & \hspace{1em} = \left(\sum_{k=2}^K\frac{(\hat{\lambda}-\hat{\lambda}^k)}{k!} \operatorname{vec} \hspace{0.1em} [\partial^k f_\theta(x)]^\top \Delta^{\otimes k} + O(\left\| \Delta \right\|_2^K)\right)^2, 
\end{aligned}
\end{equation*}
\textit{where $O(\left\| \Delta \right\|_2^K) \to 0$ as $K \to \infty $ with the input normalized such that $\left\| \Delta \right\|_2 < 1$, and the $k$-th order tensor $\partial^k f_\theta(x) \!\in\! \mathbb{R}^{d \times d \times \cdots \times d}$ is defined by:}
\begin{equation*}
    \partial^kf_\theta(x)_{t_1t_2\cdots t_k}=\frac{\partial^k}{\partial x_{t_1} \partial x_{t_2} \cdots \hspace{0.1em} \partial x_{t_k}} \hspace{0.1em} f_\theta(x),\footnote{For instance, $\partial^1 f_\theta(x)$ and $\partial^2 f_\theta(x)$ are respectively the gradient and \textit{Hessian} of $f_\theta$ evaluated at $x$.}
\end{equation*}
\textit{with which we define vectorization operation $\operatorname{vec} \hspace{0.1em} [\cdot]$ such that $\operatorname{vec} \hspace{0.1em} [\partial^{k} f_{\theta}(x)] \!\in\! \mathbb{R}^{d^{k}}$, and $a^{\otimes k} \!=\! a\otimes a\otimes\cdots\otimes a \! \in \! \mathbb{R}^{d^k}$ for any vector $a \in \mathbb{R}^{d}$, where $\otimes$ represents the Kronecker product.}

\vspace{0.5em} \noindent \textit{Proof.} We defer the proof to Appendix~\ref{app:theorem2}.
\section{Robust Alignment Adversarial Training}

Finally, we propose the \textit{\textbf{R}obust \textbf{A}lignment \textbf{A}dversarial \textbf{T}raining} (\textbf{RAAT}), which realizes the \textbf{two} aforementioned ideas for better \textit{Robust Alignment}: \uppercase\expandafter{\romannumeral1}) Reducing the perturbations applied to boundary samples for less conflict between clean and robust generalization (as discussed in Section~\ref{subsec:boundary_sample}); and \uppercase\expandafter{\romannumeral2}) Involving an additional DICAR term for the semantic alignment between input and latent spaces (as proposed in Section~\ref{subsec:DICAR}). Due to the limited space, we defer the detailed formulations to Appendix~\ref{app:RAAT}. In short, we formulate a new theoretical adversarial \textbf{risk} with the two ideas in Appendix~\ref{subapp:proposed_risk}, followed by proposing a customizable surrogate \textbf{objective} for practically optimizing this risk in Appendix~\ref{app:proposed_defense}. We end up with \textbf{two} novel AT methods, \textbf{RAAT} and \textbf{RAAT$^{++}$}, as illustrated in Table~\ref{tab:losses}, by respectively utilizing basic or more advanced loss functions in their specific implementation to optimize the surrogate objective.
% that are expected to achieve better accuracy-robustness trade-off in AT.

\section{Experiments}\label{sec:evaluation}

The effectiveness of our method is evaluated on CIFAR-10, CIFAR-100~\citep{krizhevsky2009learning}, and Tiny-ImageNet~\citep{li2015tiny}. The experiments consist of two parts. For the \textbf{first} part, based on the code implementation of Tack et al.~\citep{tack2022consistency}, we evaluate RAAT and RAAT$^{++}$ on ResNet-18~\citep{he2016deep} and WideResNet-28-10~\citep{zagoruyko2016wide}. Three typical AT baselines, PGD-AT, TRADES, and MART, along with a CR-based SOTA method, Cons-AT, are reproduced as experimental benchmarks for comparison in Section~\ref{subsec:exp_results}. In addition, a total of nine advanced related works on the accuracy-robustness trade-off problem are further included in Section~\ref{subsec:sota_compare}, where the performance reported in their previous works is adopted. This part is to provide comprehensive empirical evidence for the effectiveness and generalization ability of our method. Then, for the \textbf{second} part, by implementing our method upon the code of a current SOTA method, ReBAT~\citep{wang2024balance}, we successfully achieve the new SOTA performance beyond 11 SOTA methods on the trade-off problem with PreActResNet-18~\citep{he2016identity} architecture. This part aims to estimate the real advancement that our work can bring beyond the current level. Due to the limited space, the detailed experimental setup is provided in Appendix~\ref{subapp:sm_setup}. Three kinds of ablation studies, respectively regarding the effectiveness of the two individual ideas, the specific boundary range, and the regularization strength, are also deferred to Appendix~\ref{subapp:ablation}. The code is available at our GitHub repository.\footnote{\url{https://github.com/FlaAI/RAAT}}

\begin{table*}[h!]
\caption{Experimental results on CIFAR-10, CIFAR-100, and Tiny-ImageNet datasets with ResNet-18 and WideResNet-28-10 architectures under the $\ell_{\infty}$ threat model. All the results are averages of three runs. The final row for each dataset marked with $\uparrow$ (\%) shows the percentage improvement of the best of RAAT/RAAT$^{++}$ over the best of the four benchmarks.}\label{tab:combined}
  \centering
  \renewcommand\arraystretch{1.15}
  \definecolor{mygreen}{rgb}{0,0.6,0}
  \resizebox{\textwidth}{!}{%
  \begin{tabular}{clcccccccccc}
    \toprule
    \multirow{2.5}{*}{\textbf{Dataset}} & \multirow{2.5}{*}{\textbf{Method}} & \multicolumn{5}{c}{\textbf{ResNet-18}} & \multicolumn{5}{c}{\textbf{WideResNet-28-10}} \\
    \cmidrule(lr){3-7} \cmidrule(lr){8-12}
    & & \textbf{Clean} & \textbf{PGD-10} & \textbf{PGD-100} & \textbf{C\&W} & \textbf{AA} & \textbf{Clean} & \textbf{PGD-10} & \textbf{PGD-100} & \textbf{C\&W} & \textbf{AA} \\
    \midrule
    \multirow{7}{*}{\rotatebox{90}{\textbf{CIFAR-10}}} 
    & PGD-AT & 82.92\tiny{$\pm$0.23} & 51.81\tiny{$\pm$0.15} & 50.34\tiny{$\pm$0.14} & 48.51\tiny{$\pm$0.16} & 46.74\tiny{$\pm$0.37} & 87.49\tiny{$\pm$0.23} & 55.81\tiny{$\pm$0.33} & 54.37\tiny{$\pm$0.34} & 52.03\tiny{$\pm$0.32} & 50.96\tiny{$\pm$0.26} \\
    & TRADES & 79.67\tiny{$\pm$0.34} & 52.14\tiny{$\pm$0.26} & 51.88\tiny{$\pm$0.09} & 49.33\tiny{$\pm$0.13} & 47.62\tiny{$\pm$0.17} & 85.35\tiny{$\pm$0.57} & 57.15\tiny{$\pm$0.49} & 56.17\tiny{$\pm$0.23} & 53.61\tiny{$\pm$0.25} & 51.88\tiny{$\pm$0.18} \\
    & MART & 77.93\tiny{$\pm$0.26} & 53.61\tiny{$\pm$0.14} & 52.83\tiny{$\pm$0.30} & 48.66\tiny{$\pm$0.31} & 46.70\tiny{$\pm$0.33} & 82.78\tiny{$\pm$0.22} & 58.47\tiny{$\pm$0.28} & 57.47\tiny{$\pm$0.17} & 53.21\tiny{$\pm$0.17} & 50.89\tiny{$\pm$0.29} \\
    & Cons-AT & 83.42\tiny{$\pm$0.21} & 53.20\tiny{$\pm$0.23} & 51.68\tiny{$\pm$0.43} & 49.37\tiny{$\pm$0.25} & 47.72\tiny{$\pm$0.22} & 86.90\tiny{$\pm$0.32} & 55.71\tiny{$\pm$0.22} & 54.41\tiny{$\pm$0.19} & 52.95\tiny{$\pm$0.31} & 50.83\tiny{$\pm$0.37} \\
    & \cellcolor{black!10}\textbf{RAAT} & \cellcolor{black!10}\textbf{83.69}\tiny{$\pm$0.27} & \cellcolor{black!10}54.55\tiny{$\pm$0.14} & \cellcolor{black!10}53.50\tiny{$\pm$0.23} & \cellcolor{black!10}51.07\tiny{$\pm$0.12} & \cellcolor{black!10}\textbf{47.94}\tiny{$\pm$0.25} & \cellcolor{black!10}\textbf{87.54}\tiny{$\pm$0.16} & \cellcolor{black!10}58.22\tiny{$\pm$0.28} & \cellcolor{black!10}57.64\tiny{$\pm$0.20} & \cellcolor{black!10}54.68\tiny{$\pm$0.27} & \cellcolor{black!10}51.76\tiny{$\pm$0.24} \\
    & \cellcolor{black!10}\textbf{RAAT$^{++}$} & \cellcolor{black!10}79.70\tiny{$\pm$0.24} & \cellcolor{black!10}\textbf{54.87}\tiny{$\pm$0.16} & \cellcolor{black!10}\textbf{53.87}\tiny{$\pm$0.11} & \cellcolor{black!10}\textbf{51.64}\tiny{$\pm$0.12} & \cellcolor{black!10}47.50\tiny{$\pm$0.18} & \cellcolor{black!10}82.80\tiny{$\pm$0.25} & \cellcolor{black!10}\textbf{58.76}\tiny{$\pm$0.15} & \cellcolor{black!10}\textbf{57.88}\tiny{$\pm$0.13} & \cellcolor{black!10}\textbf{55.42}\tiny{$\pm$0.22} & \cellcolor{black!10}\textbf{52.10}\tiny{$\pm$0.31} \\
    \cmidrule(lr){2-12}
    & \hspace{0.5em}\textbf{$\uparrow$ (\%)} & \textcolor{mygreen}{+0.32\%} & \textcolor{mygreen}{+2.35\%} & \textcolor{mygreen}{+1.88\%} & \textcolor{mygreen}{+4.60\%} & \textcolor{mygreen}{+0.46\%} & \textcolor{mygreen}{+0.06\%} & \textcolor{mygreen}{+0.51\%} & \textcolor{mygreen}{+0.37\%} & \textcolor{mygreen}{+3.38\%} & \textcolor{mygreen}{+0.42\%} \\
    \midrule
    \multirow{7}{*}{\rotatebox{90}{\textbf{CIFAR-100}}} 
    & PGD-AT & 56.56\tiny{$\pm$0.28} & 29.27\tiny{$\pm$0.29} & 28.71\tiny{$\pm$0.17} & 26.62\tiny{$\pm$0.16} & 25.02\tiny{$\pm$0.19} & 59.95\tiny{$\pm$0.23} & 32.18\tiny{$\pm$0.22} & 31.21\tiny{$\pm$0.31} & 28.89\tiny{$\pm$0.26} & 27.27\tiny{$\pm$0.20} \\
    & TRADES & 55.39\tiny{$\pm$0.17} & 29.61\tiny{$\pm$0.36} & 29.28\tiny{$\pm$0.38} & 27.02\tiny{$\pm$0.34} & 24.51\tiny{$\pm$0.21} & 59.51\tiny{$\pm$0.20} & 32.51\tiny{$\pm$0.50} & 32.34\tiny{$\pm$0.26} & 29.74\tiny{$\pm$0.37} & 27.71\tiny{$\pm$0.42} \\
    & MART & 49.83\tiny{$\pm$0.13} & 30.60\tiny{$\pm$0.19} & 30.31\tiny{$\pm$0.24} & 27.27\tiny{$\pm$0.26} & 25.00\tiny{$\pm$0.22} & 56.84\tiny{$\pm$0.19} & 34.12\tiny{$\pm$0.29} & 33.70\tiny{$\pm$0.21} & 29.66\tiny{$\pm$0.25} & 27.97\tiny{$\pm$0.26} \\
    & Cons-AT & \textbf{58.53}\tiny{$\pm$0.25} & 29.99\tiny{$\pm$0.23} & 29.13\tiny{$\pm$0.26} & 26.98\tiny{$\pm$0.16} & 25.39\tiny{$\pm$0.30} & \textbf{60.84}\tiny{$\pm$0.28} & 32.48\tiny{$\pm$0.22} & 31.64\tiny{$\pm$0.22} & 29.21\tiny{$\pm$0.21} & 27.74\tiny{$\pm$0.31} \\
    & \cellcolor{black!10}\textbf{RAAT} & \cellcolor{black!10}\textbf{58.53}\tiny{$\pm$0.14} & \cellcolor{black!10}29.87\tiny{$\pm$0.20} & \cellcolor{black!10}29.24\tiny{$\pm$0.21} & \cellcolor{black!10}27.06\tiny{$\pm$0.12} & \cellcolor{black!10}\textbf{25.65}\tiny{$\pm$0.26} & \cellcolor{black!10}59.89\tiny{$\pm$0.24} & \cellcolor{black!10}33.17\tiny{$\pm$0.22} & \cellcolor{black!10}32.66\tiny{$\pm$0.15} & \cellcolor{black!10}30.84\tiny{$\pm$0.18} & \cellcolor{black!10}27.81\tiny{$\pm$0.15} \\
    & \cellcolor{black!10}\textbf{RAAT$^{++}$} & \cellcolor{black!10}53.12\tiny{$\pm$0.18} & \cellcolor{black!10}\textbf{31.43}\tiny{$\pm$0.10} & \cellcolor{black!10}\textbf{31.11}\tiny{$\pm$0.13} & \cellcolor{black!10}\textbf{28.97}\tiny{$\pm$0.16} & \cellcolor{black!10}25.32\tiny{$\pm$0.08} & \cellcolor{black!10}58.30\tiny{$\pm$0.18} & \cellcolor{black!10}\textbf{35.01}\tiny{$\pm$0.15} & \cellcolor{black!10}\textbf{34.52}\tiny{$\pm$0.17} & \cellcolor{black!10}\textbf{32.17}\tiny{$\pm$0.22} & \cellcolor{black!10}\textbf{28.52}\tiny{$\pm$0.11} \\
    \cmidrule(lr){2-12}
    & \hspace{0.5em}\textbf{$\uparrow$ (\%)} & \textcolor{Gray}{+0.00\%} & \textcolor{mygreen}{+2.71\%} & \textcolor{mygreen}{+2.62\%} & \textcolor{mygreen}{+6.23\%} & \textcolor{mygreen}{+1.02\%} & \textcolor{Salmon}{-1.40\%} & \textcolor{mygreen}{+2.62\%} & \textcolor{mygreen}{+2.43\%} & \textcolor{mygreen}{+8.17\%} & \textcolor{mygreen}{+2.56\%} \\
    \midrule
    \multirow{7}{*}{\rotatebox{90}{\textbf{Tiny-ImageNet}}} 
    & PGD-AT & 46.32\tiny{$\pm$0.12} & 21.75\tiny{$\pm$0.20} & 21.52\tiny{$\pm$0.24} & 18.79\tiny{$\pm$0.18} & 17.07\tiny{$\pm$0.17} & 47.79\tiny{$\pm$0.21} & 23.97\tiny{$\pm$0.21} & 23.59\tiny{$\pm$0.24} & 21.48\tiny{$\pm$0.12} & 20.00\tiny{$\pm$0.16} \\
    & TRADES & 46.75\tiny{$\pm$0.28} & 21.62\tiny{$\pm$0.37} & 21.52\tiny{$\pm$0.35} & 18.83\tiny{$\pm$0.25} & 16.60\tiny{$\pm$0.38} & \textbf{51.14}\tiny{$\pm$0.28} & 24.84\tiny{$\pm$0.27} & 24.58\tiny{$\pm$0.44} & 22.75\tiny{$\pm$0.26} & 20.02\tiny{$\pm$0.38} \\
    & MART & 39.70\tiny{$\pm$0.10} & \textbf{22.98}\tiny{$\pm$0.14} & 22.79\tiny{$\pm$0.09} & 19.26\tiny{$\pm$0.12} & 17.18\tiny{$\pm$0.16} & 45.57\tiny{$\pm$0.23} & 26.21\tiny{$\pm$0.22} & 25.92\tiny{$\pm$0.12} & 23.11\tiny{$\pm$0.19} & 21.07\tiny{$\pm$0.12} \\
    & Cons-AT & 46.54\tiny{$\pm$0.14} & 22.58\tiny{$\pm$0.19} & 21.70\tiny{$\pm$0.16} & 19.88\tiny{$\pm$0.14} & 17.60\tiny{$\pm$0.23} & 50.12\tiny{$\pm$0.19} & 25.05\tiny{$\pm$0.15} & 24.42\tiny{$\pm$0.25} & 22.17\tiny{$\pm$0.16} & 20.64\tiny{$\pm$0.22} \\
    & \cellcolor{black!10}\textbf{RAAT} & \cellcolor{black!10}\textbf{46.77}\tiny{$\pm$0.13} & \cellcolor{black!10}22.63\tiny{$\pm$0.14} & \cellcolor{black!10}22.28\tiny{$\pm$0.13} & \cellcolor{black!10}19.97\tiny{$\pm$0.10} & \cellcolor{black!10}\textbf{17.88}\tiny{$\pm$0.12} & \cellcolor{black!10}49.32\tiny{$\pm$0.10} & \cellcolor{black!10}25.41\tiny{$\pm$0.16} & \cellcolor{black!10}24.65\tiny{$\pm$0.13} & \cellcolor{black!10}22.28\tiny{$\pm$0.12} & \cellcolor{black!10}21.30\tiny{$\pm$0.14} \\
    & \cellcolor{black!10}\textbf{RAAT$^{++}$} & \cellcolor{black!10}41.69\tiny{$\pm$0.11} & \cellcolor{black!10}22.92\tiny{$\pm$0.11} & \cellcolor{black!10}\textbf{22.83}\tiny{$\pm$0.10} & \cellcolor{black!10}\textbf{20.62}\tiny{$\pm$0.12} & \cellcolor{black!10}17.21\tiny{$\pm$0.09} & \cellcolor{black!10}47.96\tiny{$\pm$0.14} & \cellcolor{black!10}\textbf{26.54}\tiny{$\pm$0.14} & \cellcolor{black!10}\textbf{26.01}\tiny{$\pm$0.16} & \cellcolor{black!10}\textbf{23.78}\tiny{$\pm$0.07} & \cellcolor{black!10}\textbf{22.12}\tiny{$\pm$0.08} \\
    \cmidrule(lr){2-12}
    & \hspace{0.5em}\textbf{$\uparrow$ (\%)} & \textcolor{mygreen}{+0.01\%} & \textcolor{Salmon}{-0.26\%} & \textcolor{mygreen}{+0.18\%} & \textcolor{mygreen}{+3.73\%} & \textcolor{mygreen}{+1.59\%} & \textcolor{Salmon}{-1.41\%} & \textcolor{mygreen}{+1.13\%} & \textcolor{mygreen}{+0.35\%} & \textcolor{mygreen}{+2.75\%} & \textcolor{mygreen}{+5.09\%} \\
    \bottomrule
  \end{tabular}
  }
\end{table*}

\subsection{Comparison with Four Benchmarks}\label{subsec:exp_results}

First, we report large-scale comparison results of the experimental methods regarding both adversarial robustness and clean accuracy. Specifically, white-box PGD-10, PGD-100, C\&W, and Auto-Attack (AA) with default settings and random start are adopted as the adversaries. All the reported results are averages of three runs, with the specific performance of each run acquired on the best checkpoint achieving the highest PGD-10 accuracy. The results on CIFAR-10, CIFAR-100 and Tiny-ImageNet datasets under the $\ell_{\infty}$ threat model are shown in Table~\ref{tab:combined}. Our methods achieve the best performance across ResNet-18 and WideResNet-28-10 architectures under both clean accuracy and different adversaries, significantly exceeding benchmarks, which verifies the effectiveness of the proposed defense. For clarity, we also calculate the percentage improvement of the best of RAAT/RAAT$^{++}$ over the best of the four benchmarks for each metric, dataset and architecture. Notably, due to their specific surrogate loss functions, RAAT and RAAT$^{++}$ can be viewed as natural extensions of PGD-AT and MART, respectively. When comparing RAAT with PGD-AT, and RAAT$^{++}$ with MART, it can be found that the corresponding performance on clean and adversarially perturbed data is improved together. This confirms the success of our target to harmonize clean accuracy and robustness. Furthermore, the experimental results on all three datasets with ResNet-18 architecture under the $\ell_{2}$ threat model are also demonstrated by Table~\ref{tab:l2} in Appendix~\ref{app:exp}, which shows a similar tendency to Table~\ref{tab:combined} and also supports our statements above.

\begin{figure*}[ht]
\begin{minipage}{0.37\linewidth}
\captionof{table}{Comparison of RAAT with \textbf{nine related works} on the trade-off with CIFAR-100 and ResNet-18 under $\ell_{\infty}$ threat model. Despite the related works usually employ recognized tricks (as in Section~\ref{subsec:estimate}) while our RAAT builds only upon simple PGD-AT, it still achieves the best trade-off.}
\label{tab:additional_related}
\centering
\renewcommand\arraystretch{1.4}
\resizebox{6.6cm}{!}{
    \begin{tabular}{llccc}
    \toprule
    \textbf{Method} & & \textbf{Clean} & \textbf{AA} & \textbf{Mean} \\
    \midrule
    MMA & {\scriptsize ICLR'20} \hspace{-0.3em} & 60.60 \footnotesize{$\pm$ 0.60} & 18.40 \footnotesize{$\pm$ 0.20} & 39.50 \\
    % FAT~\citep{zhang2020attacks} & {\scriptsize ICML'20} \hspace{-0.3em} & 66.74 $\pm$ 0.28 & 20.88 $\pm$ 0.13 & \underline{43.81} \\
    AWP & {\scriptsize NeurIPS'20} \hspace{-0.3em} & 55.16 \footnotesize{$\pm$ 0.27} & 25.16 \footnotesize{$\pm$ 0.39} & 40.16 \\
    GAIRAT & {\scriptsize ICLR'21} \hspace{-0.3em} & 58.43 \footnotesize{$\pm$ 0.28} & 17.54 \footnotesize{$\pm$ 0.33} & 37.99 \\
    MAIL & {\scriptsize NeurIPS'21} \hspace{-0.3em} & 60.74 \footnotesize{$\pm$ 0.15} & 22.44 \footnotesize{$\pm$ 0.53} & 41.59 \\
    TE & {\scriptsize ICLR'22} \hspace{-0.3em} & 56.45 \footnotesize{$\pm$ \hspace{1.25ex}--\hspace{1.25ex}} & 26.30 \footnotesize{$\pm$ \hspace{1.25ex}--\hspace{1.25ex}} & 41.38 \\
    HAT & {\scriptsize ICLR'22} \hspace{-0.3em} & 59.19 \footnotesize{$\pm$ 0.07} & 23.75 \footnotesize{$\pm$ 0.14} & 41.47 \\
    SOVR & {\scriptsize ICML'23} \hspace{-0.3em} & 52.10 \footnotesize{$\pm$ 0.80} & 24.30 \footnotesize{$\pm$ 0.20} & 38.20 \\
    % ReBAT & {\scriptsize NeurIPS'23} \hspace{-0.3em} & 56.13 \footnotesize{$\pm$ \hspace{1.25ex}--\hspace{1.25ex}} & 27.60 \footnotesize{$\pm$ \hspace{1.25ex}--\hspace{1.25ex}} & 41.87 \\
    ADR & {\scriptsize ICLR'24} \hspace{-0.3em} & 56.54 \footnotesize{$\pm$ \hspace{1.25ex}--\hspace{1.25ex}} & 26.42 \footnotesize{$\pm$ \hspace{1.25ex}--\hspace{1.25ex}} & 41.48 \\
    PIAT & {\scriptsize TIFS'25} \hspace{-0.3em} & 56.04 \footnotesize{$\pm$ \hspace{1.25ex}--\hspace{1.25ex}} & 26.09 \footnotesize{$\pm$ \hspace{1.25ex}--\hspace{1.25ex}} & 41.07 \\
    \midrule
    \textbf{RAAT} & {\scriptsize \textbf{Ours}} \hspace{-0.3em} & 58.53 \footnotesize{$\pm$ 0.14} & 25.65 \footnotesize{$\pm$ 0.26} & \textbf{42.09} \\
    \bottomrule
  \end{tabular}
  }
\end{minipage}\hspace{1em}
\begin{minipage}{0.6\linewidth}
\captionof{table}{Comparison of RAAT$^{\#}$ with \textbf{11 current SOTA AT methods} either in general or on the accuracy-robustness trade-off problem with PreActResNet-18 under $\ell_{\infty}$ threat model. The best results are marked in bold.}
\label{tab:SOTAs_PRN18}
\centering
\renewcommand\arraystretch{1.309}
\resizebox{10.7cm}{!}{
    \begin{tabular}{llcccccccc}
    \toprule
    \multicolumn{2}{l}{\multirow{2.3}{*}{\textbf{Method}}} & \multicolumn{4}{c}{\textbf{CIFAR-10}} & \multicolumn{4}{c}{\textbf{CIFAR-100}} \\
    \cmidrule(r){3-6} \cmidrule(r){7-10}
    \multicolumn{2}{l}{} & \textbf{Clean} & \textbf{AA} & \textbf{Mean} & \textbf{NRR} & \textbf{Clean} & \textbf{AA} & \textbf{Mean} & \textbf{NRR} \\
    \midrule
    WA & {\scriptsize UAI'18} & 83.50 & 49.89 & 66.695 & 62.461 & 57.26 & 25.83 & 41.545 & 35.601 \\
    MMA & {\scriptsize ICLR'20} & \textbf{85.50} & 37.20 & 61.350 & 51.844 & \textbf{60.60} & 18.40 & 39.500 & 28.229 \\
    AWP & {\scriptsize NeurIPS'20} & 81.11 & 50.09 & 65.600 & 61.933 & 54.10 & 25.16 & 39.630 & 34.347 \\
    GAIRAT & {\scriptsize ICLR'21} & 78.70 & 37.70 & 58.200 & 50.979 & 52.00 & 19.80 & 35.900 & 28.680 \\
    KD+SWA & {\scriptsize ICLR'21} & 84.06 & 49.82 & 66.940 & 62.562 & 57.17 & 25.66 & 41.415 & 35.422 \\
    EWAT & {\scriptsize ICML'21} & 82.80 & 48.20 & 65.500 & 60.931 & 54.20 & 23.52 & 38.860 & 32.805 \\
    MAIL & {\scriptsize NeurIPS'21} & 79.50 & 39.60 & 59.550 & 52.867 & 46.50 & 16.70 & 31.600 & 24.574 \\
    TE & {\scriptsize ICLR'22} & 82.04 & 50.12 & 66.080 & 62.225 & 56.41 & 25.84 & 41.125 & 35.444 \\
    SOVR & {\scriptsize ICML'23} & 81.90 & 49.40 & 65.650 & 61.628 & 52.10 & 24.30 & 38.200 & 33.142 \\
    ReBAT & {\scriptsize NeurIPS'23} & 82.09 & 50.72 & 66.405 & 62.700 & 56.13 & 27.60 & 41.865 & 37.004 \\
    RPAT & {\scriptsize ICCV'25} & 82.27 & 50.75 & 66.510 & 62.776 & 56.43 & 27.63 & 42.030 & 37.096 \\
    \midrule
    \textbf{RAAT$^{\#}$} & {\scriptsize \textbf{Ours}} & 82.76 & \textbf{51.65} & \textbf{67.205} & \textbf{63.605} & 56.92 & \textbf{28.08} & \textbf{42.500} & \textbf{37.607} \\
     \bottomrule
    \end{tabular}
  }
\end{minipage}
\end{figure*}

\subsection{Comparison with Nine Related Works}\label{subsec:sota_compare}

In this section, we compare the proposed RAAT with nine advanced related works detailed in Appendix~\ref{subapp:backg_tradeoff}. All of these works are from top-tier conferences or journals in the past five years, effectively representing the current research status in the accuracy-robustness trade-off problem of AT. The comparison is based on CIFAR-100 dataset, and there are mainly two reasons for this choice: \uppercase\expandafter{\romannumeral1}) This is expected to better fit larger-scale real-world practices, and \uppercase\expandafter{\romannumeral2}) A recent work~\citep{geiping2022stochastic} worries that CIFAR-10 might not be representative enough for SGD training, which is the default setting for AT including ours. Of course, considering these two points, Tiny-ImageNet dataset would also be a reasonable option, but unfortunately, most related works simply do not report any results on it. The records of these related works come from either their original papers or \citet{liu2021probabilistic}.
% \footnote{Notably, the record for ReBAT is from not ResNet-18 but PreActResNet-18 architecture, as it has not been tested on ResNet-18 in the original paper and previous works. As PreActResNet-18 is more powerful, the scores for ReBAT may be slightly higher, which means the actual improvement of our RAAT beyond ReBAT is expected to be more significant than the results here.}
We report the mean of clean accuracy and robust score under AA as the measure of the trade-off performance. As illustrated in Table~\ref{tab:additional_related}, even our basic RAAT method can outperform all these related works in the trade-off problem, which further confirms the effectiveness of our work in harmonizing clean accuracy and robustness in AT.

\subsection{Estimation of Advancement beyond 11 SOTAs}\label{subsec:estimate}

Different from the simple baselines we reproduced in Section~\ref{subsec:exp_results}, most SOTA methods include multiple ingenious tricks that are widely recognized as effective. For instance, ReBAT employs WA~\citep{izmailov2018averaging}, averaging multiple points along the SGD trajectory for better generalization, through the exponential moving average (EMA)~\citep{rebuffi2021data} strategy with a decay rate of 0.999. Thus, directly comparing the results of our method built upon simple PGD-AT with those SOTAs enhanced by tricks, like in Section~\ref{subsec:sota_compare}, is not completely fair for us. Therefore, to fairly estimate the real advancement that our method contributes, the most straightforward approach is to implement it upon one of the current SOTA frameworks. In this section, we implement the mechanism of the proposed RAAT upon the code of ReBAT~\citep{wang2024balance}, a current SOTA AT method on the accuracy-robustness trade-off problem. We refer to this variant as RAAT$^{\#}$, and compare it with WA itself (\ie, with PGD-AT), KD+SWA~\citep{chen2021robust} which shares a similar idea, and nine existing SOTAs introduced in Appendix~\ref{subapp:backg_tradeoff} on PreActResNet-18, as shown in Table~\ref{tab:SOTAs_PRN18}. The clean and AA scores of the SOTAs are from the best checkpoint among their original papers, Wang et al.~\citep{wang2025failure}, and our reproduction if any. Beyond the mean score, we further introduce the Natural-Robustness Ratio (NRR)~\citep{gowda2024conserve}:
\begin{equation*}
\textit{NRR} = \frac{2 \times \textit{Clean Accuracy} \times \textit{Adversarial Robustness}}{\textit{Clean Accuracy} + \textit{Adversarial Robustness}},
\end{equation*}
which emphasizes how well a model strikes the balance between accuracy and robustness, as an additional metric for the trade-off problem. Using PreActResNet-18 here is because, expanding the scope of the experimental model is beneficial, and on the other hand, compared to the previous two, this model is more commonly included in SOTA works. The results show that our RAAT$^{\#}$ successfully achieves new SOTA performance beyond previous SOTAs, which strongly supports the contribution of our work.

\section{Conclusion}\label{sec:conclusion}

In this work, we study how to harmonize clean accuracy and robustness in AT given the current trade-off between them. To the best of our knowledge, this is the \textbf{first} work explicitly attributing this trade-off problem to the \textbf{misalignment of input and latent spaces}. In response, we define a new AT target, \textbf{\textit{Robust Alignment}}, which involves \textbf{two} novel ideas to mitigate such misalignment: \uppercase\expandafter{\romannumeral1}) Reducing the perturbation intensity for boundary samples to facilitate the learning of perturbations, and \uppercase\expandafter{\romannumeral2}) Adding the newly proposed DICAR term to encourage the alignment between input and latent representations. Accordingly, two surrogate methods, RAAT and RAAT$^{++}$, are proposed for better accuracy-robustness trade-off, outperforming four baselines and nine related works. Furthermore, RAAT$^{\#}$, an advanced implementation of RAAT, builds the new SOTA beyond 11 previous SOTA works. We expect that both the insensitivity of boundary samples to perturbation intensity and the definition of \textit{Robust Alignment} to align input and latent spaces would be found insightful and help update the current knowledge of AT. Looking ahead, exploring more effective strategies to better approach \textit{Robust Alignment} might be a promising direction for future works, especially those improving the computational efficiency, which remains another critical concern for existing AT methods. 

\section*{Acknowledgements}

This work was supported by the Ministry of Science and Technology of the People’s Republic of China (Grant No:  2025YFE0200100), the National Natural Science Foundation of China (Grant No:  62372122), the Research Grants Council (Grant No: 15224124 and 15207725), Hong Kong SAR, China.

{
    \small
    \bibliographystyle{ieeenat_fullname}
    \bibliography{main}
}

\clearpage
\onecolumn

\appendix

\section*{\centering{\Large Appendix}}

% {\centering{\large Robust Alignment: Harmonizing Clean Accuracy and Adversarial Robustness in Adversarial Training \\}}

\section*{Table of Contents}

% \stopcontents[main]
\startcontents[app] 
\begin{mdframed}[
    topline=true,bottomline=true,
    leftline=false,rightline=false,
    linewidth=1pt,
    innertopmargin=.3\baselineskip,
    innerbottommargin=1.2\baselineskip,
    skipabove=\baselineskip,skipbelow=\baselineskip]
  \printcontents[app]{l}{1}{}
\end{mdframed}

\section{Detailed Background and Related Works}

\subsection{Adversarial Robustness}

Adversarial robustness indicates the vulnerability of DNN classifiers under intended perturbations by an adversary, which is commonly measured with the test accuracy under adversarial attacks~\citep{bai2021recent}. Since \citet{szegedy2014intriguing} introduced the concept of adversarial attack, many effective attack methods have been proposed. For instance, the Fast Gradient Sign Method (FGSM)~\citep{goodfellow2015explaining} implements the perturbation according to the gradient of the loss function \textit{w.r.t.} the input sample. The Projected Gradient Descent (PGD) attack~\citep{madry2018towards} can be viewed as a variant of FGSM. It generates the perturbation by FGSM iteratively and then projects it to the $\epsilon$-ball of the input sample. The C\&W attack~\citep{carlini2017towards} no longer relies on $\epsilon$-ball as a constraint of perturbation radius, but formulates a regularization term leading to small perturbation instead. Auto-Attack (AA)~\citep{croce2020reliable} forms a parameter-free and user-independent ensemble of attacks for frequent pitfalls in practice like improper tuning of hyper-parameters and gradient obfuscation or masking. Due to their effectiveness, PGD, C\&W and AA are commonly used to evaluate the adversarial robustness of DNNs.

\vspace{1ex} \noindent With the development of adversarial attack, there are also a number of defense methods proposed to improve the adversarial robustness of DNNs, including Defense Distillation~\citep{papernot2017practical}, Feature Squeezing~\citep{xu2017feature}, Input Denoising~\citep{guo2018countering,liao2018defense} and Randomization~\citep{xie2018mitigating}. However, most of these methods have been proven subsequently to rely on obfuscated gradients~\citep{athalye2018obfuscated} and be ineffective against advanced adaptive attacks~\citep{tramer2020adaptive}.

\subsection{Adversarial Training}

Currently, Adversarial Training~\citep{goodfellow2015explaining,madry2018towards} is widely recognized as the most effective and practical method to acquire adversarially robust DNNs~\citep{athalye2018obfuscated,dong2020benchmarking}. Different from clean training, in the context of AT, the model is directly trained on adversarially augmented samples instead of only natural ones. Specifically, the optimization objective of AT can be defined as a $\min$-$\max$ problem~\citep{madry2018towards}:
\begin{equation}\label{eq:AT}
    \min_{\bm{\theta}} \frac{1}{n} \hspace{0.1em} \sum_{i=1}^n \max_{\hspace{0.1em} \Vert \mathbf{x}_i' - \mathbf{x}_i \Vert_{p} \leq \epsilon \hspace{0.1em}} \mathcal{L}(f_{\bm{\theta}}(\mathbf{x}_i'), y_i),
\end{equation}
where $n$ is the number of training samples and $\mathbf{x}_i'$ is the adversarial sample generated by adding the strongest perturbation to the natural one $\mathbf{x}_i$ within the $\epsilon$-ball under $L_p$-norm distance, which can be used by the outer minimization of classification loss $\mathcal{L}$ to train a robust model. Research has shown that this is equivalent to optimizing an upper bound of natural risk on the original data, which implies that AT can serve as a principled defense against adversarial attacks~\citep{tao2021better}.

\vspace{1ex} \noindent PGD-AT~\citep{madry2018towards} is the first method demonstrated to be effective for solving this $\min$-$\max$ problem and training moderately robust DNNs~\citep{athalye2018obfuscated,wang2020improving}. Based on \textit{Danskin’s Theorem}~\citep{danskin2012theory}, PGD-AT proposed to find a constrained maximizer of the inner maximization by PGD, which is believed sufficiently close to the optimal attack, and then use the maximizer as an actual data point for the outer minimization through gradient descent. Another typical AT method, TRADES~\citep{zhang2019theoretically}, proposed to decompose the robust error into natural error and boundary error to balance the trade-off between natural accuracy and robustness. Specifically, boundary error occurs when the specific data point is sufficiently close to the decision boundary that can easily cross it under slight perturbation. This is also believed as one reason for the existence of adversarial samples~\citep{bai2021recent,wang2022tsfool}. One problem of TRADES is that the boundary error is designed to push each pair of benign and adversarial samples together, no matter whether the benign data are classified correctly or not~\citep{bai2021recent}. As a follow-up work, MART~\citep{wang2020improving} further investigates the influence of correctly classified and misclassified samples for adversarial robustness separately, and suggests applying robust error uniformly, plus boundary error only on misclassified samples. 
% The specific optimization objectives of the aforementioned AT methods are illustrated in Table~\ref{tab:losses}.

\subsection{Accuracy-Robustness Trade-Off in AT}\label{subapp:backg_tradeoff}

One of the most important flaws of AT is an inherent trade-off between clean accuracy and adversarial robustness~\citep{tsipras2019robustness,zhang2019theoretically,raghunathan2019adversarial,raghunathan2020understanding,wang2020once,bai2021recent,rade2022reducing,yin2023push,liu2025parameter}. Specifically, the improvement in robustness achieved by AT is always at the cost of the reduction in model accuracy compared with standard training. As a consequence, AT inevitably degrades user experience under benign environments, which seriously hinders its real-world application. As shown in Figure~\ref{fig:AT_generalization}, a widely recognized cause of this problem is that, to achieve predictive robustness under adversarial perturbation within the $\epsilon$-ball of each natural data point uniformly, AT tends to learn a more complicated decision boundary than standard training~\citep{dong2022exploring,rade2022reducing,cheng2022cat,yang2022one,yin2023push,liu2025parameter}, which harms the generalization ability of the model on unseen data. 

\vspace{1ex} \noindent In response, it has recently become an active research area to variously craft or weight data points near the decision boundary for a better trade-off between accuracy and robustness. However, to date, there are still mixed views on the role and principle of these boundary samples in AT. \textbf{One side} believes the data closer to the current decision boundaries should be regarded as more critical and learned with larger weights or enhanced patterns. Representatively, GAIRAT~\citep{zhang2021geometry} proposed that a natural data point closer to (or farther from) the class boundary is less (or more) robust, and the corresponding adversarial data point should be assigned with larger (or smaller) weight. MAIL~\citep{liu2021probabilistic} further revealed the unreliability of existing measures of the closeness, and proposed three types of probabilistic margin for measuring the closeness and reweighting adversarial data. \textbf{The other side} argues that reducing perturbations or weights for those data points near the decision boundary can bring significant benefits to AT generalization. For instance, MMA~\citep{ding2020mma} proposed to use adaptive $\epsilon$ for adversarial perturbations to directly estimate and maximize the margin between data and the decision boundary. HAT~\citep{rade2022reducing} has a similar idea, which is realized by creating artificial helper-examples. TE~\citep{dong2022exploring} suggested that one-hot labels might be noisy for the boundary samples because they naturally lie close to the decision boundary, which makes it essentially difficult to assign high-confident one-hot labels for all perturbed samples within the $\epsilon$-ball of them~\citep{stutz2020confidence,cheng2022cat}. So the model may try to memorize these hard samples during AT, leading to a more complicated decision boundary. All in all, with this inconclusive controversy, how to appropriately treat data points near the decision boundary is still an open question in AT.

\vspace{1ex} \noindent Besides, some previous works also study the trade-off problem from other perspectives. For instance, AWP~\citep{wu2020adversarial} proposed a double perturbation mechanism that can flatten the loss landscape by weight perturbation to improve robust generalization. EWAT~\citep{kim2021entropy} weighs the loss for each adversarial sample proportionally to the entropy of its prediction distribution during AT to focus on those with more uncertain labels. SOVR~\citep{kanai2023one} proposed to increase logit margins of important samples by switching from cross-entropy to a new one-vs-therest loss for a considerable trade-off between accuracy and robustness. ADR~\citep{wu2024annealing} generates soft labels as a better guidance mechanism that accurately reflects the distribution shift under attack during AT. ReBAT~\citep{wang2024balance} views AT as a dynamic mini-max game between the model trainer and the attacker, and proposes to rebalance the two players by either regularizing the trainer's capacity or improving the attack strength. PIAT~\citep{liu2025parameter} tunes the model parameters by interpolating them from the previous and current epochs, reducing the oscillations during the training process and moderating the change of decision boundaries. RPAT~\citep{wang2025failure} suggests that the over-sufficient learning of hard adversarial samples degrades the decision boundary and contributes to the trade-off problem, and thus release it by encouraging the model perception to change smoothly with input perturbations.

\vspace{1ex} \noindent At the same time, there are also some works theoretically doubting the assumption that adversarial robustness should be inherently at the cost of clean accuracy~\citep{stutz2019disentangling,yang2020closer,bai2021recent}. Representatively, \citet{yang2020closer} found that various common datasets are separated in class distribution with the separation larger than 2$\epsilon$ in usual, which directly indicates the existence of both robust and accurate classifiers and implies that the current trade-off problem is just an undesirable result caused by the existing AT methods themselves. Inspired by this, different from the previous works that aim at a better trade-off between accuracy and robustness, the vision of this work is to directly harmonize them, such that the clean and robust accuracy scores can be improved concurrently.

\subsection{AT with Consistency Regularization}\label{subapp:backg_CR}

Consistency Regularization (CR) aims to encourage a model to produce invariant representation for different variants of the same sample~\citep{fan2023revisiting}. This is based on an assumption that slight perturbations like randomness within DNNs (\eg, with Dropout) or data augmentation transformations should not modify the model prediction of the same input~\citep{zhang2022alleviating}. Typically, this idea can be realized through an additional consistency regularization term appended to the loss function, which has been widely adopted in existing Semi-Supervised Learning (SSL) algorithms~\citep{fan2023revisiting}, such as $\Pi$-model~\citep{sajjadi2016regularization,laine2017temporal}, Temporal Ensembling~\citep{laine2017temporal}, Mean Teacher~\citep{tarvainen2017mean} and Interpolation Consistency Training~\citep{verma2022interpolation}. 

\vspace{1ex} \noindent Recently, it has been found that there is an implicit connection between the goals of CR and AT~\citep{zhang2022alleviating}. Specifically, CR forces the model to give the same output distribution when the input or parameters are slightly perturbed, which covers the aim of AT when the perturbation is generated adversarially. Besides, as adversarial robustness essentially refers to model stability around naturally occurring inputs, learning to satisfy such a constraint should not inherently require labels~\citep{carmon2019unlabeled}, which provides the theoretical basis for the extension of CR, as an unsupervised mechanism, in the field of AT. Therefore, there have been some works exploring how to make use of CR ideas to improve adversarial robustness. 

\vspace{1ex} \noindent One of the most straightforward ideas is to introduce additional unlabeled data and their adversarial augmentations for consistency regularization~\citep{alayrac2019labels,carmon2019unlabeled,najafi2019robustness,zhai2019adversarially,li2022semi}, which can be viewed as a natural extension of the SSL algorithm $\Pi$-model in the context of robustness. However, these works have to rely on additional data, which not only brings extra costs in data collection and computation but also violates the conventional AT settings~\citep{pang2021bag}. Subsequent works focus more on directly introducing the principle of CR into the existing AT framework. For instance, TE~\citep{dong2022exploring} makes use of the principle of Temporal Ensembling to maintain an ensemble prediction of each adversarial example and penalize the difference between the current prediction and the ensemble prediction. This is to regularize the predictions of adversarial examples from being over-confident, which is expected to reduce the impact of label noise during AT. \citet{zhang2022alleviating} integrates the strategy from Mean Teacher into AT to smooth the model and reduce possible overfitting. Specifically, it encourages the prediction distribution of a student model over adversarial examples to be consistent with that of a teacher model over clean samples. \citet{liu2022mutual} and \citet{kuang2024defense} share similar ideas in introducing teacher-student learning strategies. Still, these methods basically transfer the existing CR methods to the AT task. Recently, Cons-AT~\citep{tack2022consistency} proposed a new consistency target directly from the perspective of robustness that the predictive distributions after attacking from two different augmentations of the same instance should be similar with each other. The underlying principle is that the most confusing class, and further, the most frequent attack direction, of specific samples is a kind of intrinsic information belonging to the so-called ``dark'' knowledge~\citep{hinton2014distilling} and should be consistent over its different augmented variants. 

\vspace{1ex} \noindent Despite these previous efforts, the principle of CR in the context of AT remains highly unexplored, especially \textit{w.r.t.} its potential on the accuracy-robustness trade-off problem. Specifically, these previous works mainly focus on the ability of CR to alleviate robust overfitting, with various theoretical explanations raised from different perspectives, such as model smoothness~\citep{chen2021robust} and the flatness of the weight loss landscape~\citep{wu2020adversarial,stutz2021relating}.\footnote{Robust overfitting refers to a phenomenon that, at certain epochs during AT (\eg, after the first learning rate decay), model robustness will drop sharply, resulting in a significant gap in robust accuracy score between adversarially perturbed training and test data~\citep{rice2020overfitting,jia2024revisiting}.} Predictably, the empirical results also show that, although contributing significantly smaller robust generalization gaps in the last epochs of AT, these works only slightly improve robustness at the best checkpoint of the model. What's worse, usually the price is to further hurt clean accuracy~\citep{dong2022exploring,zhang2022alleviating}. The only aforementioned works taking the accuracy-robustness trade-off into consideration are the ones based on the $\Pi$-model~\citep{schmidt2018adversarially,alayrac2019labels,carmon2019unlabeled}. They study the trade-off problem from the perspective of sample complexity, but again, merely under a relaxed assumption with extra data. All in all, to date, whether and how CR can benefit the accuracy-robustness trade-off in typical AT without extra data remains to be further studied.

\section{Proof of Theorems}

\subsection{Proof of Theorem 1}\label{app:theorem1}

Given the Gaussian model as in Definition 1, as a direct consequence of Gaussian concentration, \citet{schmidt2018adversarially} prove the following theorem, suggesting that we can learn a linear classifier achieving fixed and arbitrary classification error even with only one labeled sample.

\vspace{1ex} \noindent \textbf{Theorem 3} (Theorem 4 in \citet{schmidt2018adversarially})\textbf{.} \hspace{1pt} \textit{Let $(x, y)$ be drawn from a $(\theta^*, \sigma)$-Gaussian model with $\|\theta^*\|_2 \!=\! \sqrt{d}$ and $\sigma \!\leq\! c \cdot d^{1\!/4}$ where $c$ is a universal constant. Let $\widehat{w} \in \mathbb{R}^d$ be the vector $\widehat{w} = y \cdot x$. Then with high probability, the linear classifier $f_{\widehat{w}}$ has classification error at most 1\%.}

\vspace{1.5ex} \noindent However, the following theorem shows that, to achieve the same robust generalization error under $\ell_\infty$ perturbation, any algorithm requires more labeled samples.

\vspace{1.5ex} \noindent \textbf{Theorem 4} (Theorem 6 in \citet{schmidt2018adversarially})\textbf{.} \hspace{1pt} \textit{Let $g_n$ be any learning algorithm, \ie, a function from $n$ samples to a binary classifier $f_n$. Moreover, let $\sigma = c_1 \cdot d^{1/4}$, $\epsilon \geq 0$ and $\theta \in \mathbb{R}^d$ be drawn from $\mathcal{N}(0, I)$. We also draw $n$ samples from the $(\theta, \sigma)$-Gaussian model. Then the expected $\ell_\infty^\epsilon$ robust error of $f_n$ is at least $\frac{(1 - 1/d)}{2}$ if}
\begin{equation}
n\:\leq\:c_2\frac{\epsilon^2\:\sqrt{d}}{\log d}\:,
\end{equation}
\textit{where $c_1$ and $c_2$ are two universal constants.}

\vspace{1.5ex} \noindent Combining Theorem 3 and Theorem 4, we can see the sample complexity for robust generalization is greater than that of standard generalization by $\sqrt{d} / \log d$. As a direct corollary, given the same number of training samples, the expected gap between standard and robust generalization error can also be estimated by this.

\vspace{1.5ex} \noindent \textbf{Corollary 1.} \textit{Considering a $(\theta, \sigma)$-Gaussian model with $\|\theta\|_2 \!=\! \sqrt{d}$ and $\sigma \!\leq\! c_1 \cdot d^{1/4}$, we have an expected gap between the lower bounds of standard and robust generalization error increasing with $O(\sqrt{d}/\log d)$.}

\vspace{1.5ex} \noindent On the other hand, previous works have found that the sample complexity of robust generalization can be dramatically reduced by replacing labeled examples with additional unlabeled samples~\citep{alayrac2019labels,carmon2019unlabeled,najafi2019robustness,zhai2019adversarially}. Typically, \citet{alayrac2019labels} derive the following theorem, suggesting the specific number of additional samples needed.

\vspace{1.5ex} \noindent \textbf{Theorem 5} (Theorem 1 and Corollary 11 in \citet{alayrac2019labels})\textbf{.} \hspace{1pt} \textit{Consider the $(\theta^*, \sigma)$-Gaussian model with $\|\theta^*\|_2 = \sqrt{d}$ and $\sigma \leq \frac{1}{32}d^{1/4}$ as in \citet{schmidt2018adversarially}. Let $\widehat{w}$ be the UAT-FT estimator as in \citet{alayrac2019labels}. Then with high probability, for n=1, the linear classifier $f_{\hat{w}}$ has $\ell_\infty^\epsilon$ robust classification error at most 1\% if}
\begin{equation}
m \geq c \epsilon^2 \sqrt{d},
\end{equation}
\textit{where $c$ is a fixed, universal constant.}
% Let the UAT-FT estimator $\widehat{w} \in \mathbb{R}^d$ be the unit vector in the direction of the sample mean $\overline{z}= \frac 1m\sum _{i= 1}^{m}\hat{y} _ix_i, where$
% $\textit{ Let }\hat{w} _{sup}= y_0x_0. \textit{ Let }\overline {z}\in \mathbb{R} ^d\textit{be }$  $\overline {z}= \frac 1m\sum _{i= 1}^{m}\hat{y} _ix_i, where$ $\hat{y} _i$ = $f_{\hat{w} _{sup}}( x_i) .$ Let the $$ $$ $be$ the 
% $\overline z,i.e,\hat{w}=\overline{z}/\|\overline{z}\|_2.$ 

\vspace{1.5ex} \noindent As explained in Definition 2, Gaussian CR is a natural extension of Gaussian UAT-FT under the conventional fairness assumption of AT. The main reason for proposing Gaussian CR in this work is to highlight it is practically approximated in a completely different manner, though it follows the same theoretical principle as Gaussian UAT-FT. Therefore, along with Theorem 3, we have Corollary 2 by simply adopting Gaussian CR estimator in Theorem 5.

\vspace{1.5ex} \noindent \textbf{Corollary 2.} \textit{With the Gaussian CR estimator $\tilde{\theta}$ as in Definition 2, the generalization gap between standard and robust classifications can be filled at the cost of $O(\sqrt{d})$ additional unlabeled data points.}

\vspace{1.5ex} \noindent Finally, through combining Corollary 1 and Corollary 2, we have Theorem 1 as in the main body.

\subsection{Proof of Theorem 2}\label{app:theorem2}

As introduced in Section 3, ICT~\citep{verma2022interpolation} is a state-of-the-art method in the field of SSL. The following theorem in its original paper suggests that its CR term for unlabeled data can act as a regularizer on higher-order derivatives. 

\vspace{1.5ex} \noindent \textbf{Theorem 6} (Theorem 1 in \citet{verma2022interpolation})\textbf{.} \hspace{1pt} \textit{Let $u, u^\prime \in \mathbb{R}^d$ and $f_\theta$ be real analytic, and define $\Delta = u^{\prime} - u$. Then, for any $K \in \mathbb{N}^{+}$, there exists a pair $(\zeta, \zeta^{\prime} \in [0, \hat{\lambda}] \times [0,1])$ such that}
\begin{equation}
% \begin{aligned}
    \ell(f_{\theta}(\operatorname*{Mix}_{\lambda}(u,u^{\prime})),\operatorname*{Mix}_{\lambda}(f_{\theta}(u),f_{\theta}(u^{\prime}))) = \left(\sum_{k=2}^K\frac{(\hat{\lambda}-\hat{\lambda}^k)}{k!}\operatorname{vec}[\partial^kf_\theta(u)]^\top\Delta^{\otimes k}+E_\theta(u,u^{\prime},\lambda)\right)^2,
% \end{aligned}
\end{equation}
 
\noindent \textit{where $E_{\theta }(u, u^{\prime }, \lambda) = \frac{1}{K!}((1-\zeta^{\prime})^{K}\operatorname{vec}[\partial^{K+1}f_{\theta}(u+\zeta^{\prime}\Delta)]-\hat{\lambda}(\hat{\lambda}-\zeta)^{K}\operatorname{vec}[\partial^{K+1}f_{\theta}(u+\zeta\Delta)])^{\top}\Delta^{\otimes K} = O(\| \Delta \| _{2}^{K})$ and $\operatorname*{Mix}_{\lambda}(a, b) = \lambda \cdot a + (1 - \lambda) \cdot b$. With the input normalized such that $\left\| \Delta \right\|_2 < 1$, $O(\left\| \Delta \right\|_2^K) \to 0$ as $K \to \infty $.}
 
\vspace{2ex} \noindent In this work, given $\dot{x}$ and $\ddot{x}$ as in Definition 3, let $u = \dot{x}$ and $u' = \ddot{x}$ in Theorem 7, we have:
\begin{equation}
\ell \hspace{0.1em} \Big( f_\theta(\lambda \dot{x} + \hat{\lambda} \ddot{x}) \hspace{0.1em} \big( \lambda f_\theta(\dot{x}) + \hat{\lambda} f_\theta(\ddot{x}) \big) \Big) = \left(\sum_{k=2}^K\frac{(\hat{\lambda}-\hat{\lambda}^k)}{k!}\operatorname{vec}[\partial^kf_\theta(\dot{x})]^\top\Delta^{\otimes k} + O(\| \Delta \| _{2}^{K}) \right)^2.
\end{equation}

\noindent Further, to transfer this theorem to the context of AT, we introduce a limitation of a sufficiently small perturbation budget $\epsilon \to 0$ such that $\dot{x}' \to \dot{x}$ and $\ddot{x}' \to \ddot{x}$. Considering that $\dot{x}$ is augmented from $x$ together, we have:
\begin{equation}
\ell \hspace{0.1em} \Big( f_\theta(\lambda \dot{x}' + \hat{\lambda} \ddot{x}') \hspace{0.1em} \big( \lambda f_\theta(\dot{x}) + \hat{\lambda} f_\theta(\ddot{x}) \big) \Big) = \left(\sum_{k=2}^K\frac{(\hat{\lambda}-\hat{\lambda}^k)}{k!}\operatorname{vec}[\partial^kf_\theta(x)]^\top\Delta^{\otimes k} + O(\| \Delta \| _{2}^{K}) \right)^2.
\end{equation}

\noindent Then with the $\operatorname{vec} \hspace{0.1em} [\partial^{k} f_{\theta}(x)] \!\in\! \mathbb{R}^{d^{k}}$ defined as in \citet{verma2022interpolation}, we have Theorem 2 in the main body.

\section{Robust Alignment Adversarial Training: Overall Risk, Objective, and Algorithms}\label{app:RAAT}

In this section, we first formulate a new theoretical adversarial \textbf{risk} with the two main ideas proposed in the main text, and then propose a customizable surrogate \textbf{objective} for the practical optimization of this risk. Finally, under such a new surrogate objective, we suggest two specific methods that are expected to achieve a better accuracy-robustness trade-off in AT.

\subsection{Constructing New Adversarial Risk for Robust Alignment}\label{subapp:proposed_risk}

In this section, we formally define the proposed adversarial risk. For a $K$-class classification task ($K \geq 2$), given a dataset $\mathcal{S} = \{(\mathbf{x}_i, y_i)\}_{i = 1, ..., n}$ with $\mathbf{x}_i \in \mathbb{R}^{d}$ and $y_i \in \{1, ..., K\}$ respectively denoting a natural sample and its supervised label, as well as a DNN model $f_{\bm{\theta}}$ to predicts the input sample as $f_{\bm{\theta}}(\mathbf{x}_i) = \argmax_{k = 1, ..., K} \hspace{0.1em} \mathbf{p}_{k} (\mathbf{x}_i, \bm{\theta})$, where $\mathbf{p}_{k}$ is the probability (\ie, \textit{softmax}) of class $k$ in the prediction of $\mathbf{x}_i$, the standard adversarial risk~\citep{madry2018towards} with respect to the \textit{0-1 loss}~\citep{zhang2019theoretically} can be defined as follow:
\begin{equation}\label{eq:adv_risk_standard}
    \mathcal{R}(f_{\bm{\theta}}) = \frac{1}{n} \hspace{0.1em} \sum_{i=1}^n \max_{\hspace{0.1em} \Vert \mathbf{x}_i' - \mathbf{x}_i \Vert_{p} \leq \epsilon \hspace{0.1em}} \mathds{1}(f_{\bm{\theta}}(\mathbf{x}_i') \neq y_i) = \frac{1}{n} \hspace{0.1em} \sum_{i=1}^n \mathds{1}(f_{\bm{\theta}}(\mathbf{\hat{x}}_i') \neq y_i),
\end{equation}
where $\mathds{1}(\cdot)$ is the indicator function and $\mathbf{\hat{x}}_i'$ is the adversarial sample generated through $\mathbf{\hat{x}}_i' = \argmax_{\Vert \mathbf{x}_i' - \mathbf{x}_i \Vert_{p} \leq \epsilon} $ $\mathds{1}(f_{\bm{\theta}}(\mathbf{x}_i') \neq y_i)$. As this risk is defined on perturbed samples within the $\epsilon$-ball of all natural samples no matter whether they are correctly classified or not, MART~\citep{wang2020improving} proposed to reformulate the adversarial risk by dividing natural samples into the correctly classified subset $\mathcal{S}_{f_{\bm{\theta}}}^{+} = \{ \mathbf{x}_i : \mathbf{x}_i \in \mathcal{S}, \hspace{0.1em} f_{\bm{\theta}}(\mathbf{x}_i) = y_i \}$ and the misclassified subset $\mathcal{S}_{f_{\bm{\theta}}}^{-} = \{ \mathbf{x}_i : \mathbf{x}_i \in \mathcal{S}, \hspace{0.1em} f_{\bm{\theta}}(\mathbf{x}_i) \neq y_i \}$ with respect to the current prediction of $f_{\bm{\theta}}$, and then defining the adversarial risk separately for them.

\noindent In this work, to utilize the phenomenon observed regarding the ``boundary samples'', we further propose to divide the correctly classified samples $\mathcal{S}_{f_{\bm{\theta}}}^{+}$ into the ones close to the classification boundary as $\mathcal{S}_{f_{\bm{\theta}}}^{\circ}$ and the ones away from that as $\mathcal{S}_{f_{\bm{\theta}}}^{\bullet}$. Formally, given a natural sample $\mathbf{x}_i \in \mathcal{S}_{f_{\bm{\theta}}}^{+}$ and the perturbation radius $\epsilon$ for adversarial sample $\mathbf{\hat{x}}_i'$, a \textit{reduced adversarial sample} $\mathbf{\hat{x}}_i''$ can be denoted as:
\begin{equation}\label{eq:weak_adv_sample}
    \mathbf{\hat{x}}_i'' = \argmax_{\Vert \mathbf{x}_i'' - \mathbf{x}_i \Vert_{p} \leq \eta \cdot \epsilon} \mathds{1}(f_{\bm{\theta}}(\mathbf{x}_i'') \neq y_i),
\end{equation}
where $\eta \in (0, 1)$ is a pre-defined hyper-parameter to reduce the perturbation. Then if $\exists \hspace{0.2em} \mathbf{\hat{x}}_i''$, s.t. $f_{\bm{\theta}}(\mathbf{\hat{x}}_i'') \neq y_i$, the natural sample $\mathbf{x}_i$ can be defined as a \textit{boundary sample}, otherwise a \textit{non-boundary sample}. Accordingly, we can define $\mathcal{S}_{f_{\bm{\theta}}}^{\bullet}$ and $\mathcal{S}_{f_{\bm{\theta}}}^{\circ}$ respectively as:
\begin{align}\label{eq:S_divide_2}
    \mathcal{S}_{f_{\bm{\theta}}}^{\bullet} = \{ \mathbf{x}_i : \mathbf{x}_i \in \mathcal{S}_{f_{\bm{\theta}}}^{+}, \hspace{0.2em} f_{\bm{\theta}}(\mathbf{\hat{x}}_i'') = y_i \}, \hspace{1em} \mathcal{S}_{f_{\bm{\theta}}}^{\circ} = \{ \mathbf{x}_i : \mathbf{x}_i \in \mathcal{S}_{f_{\bm{\theta}}}^{+}, \hspace{0.2em} f_{\bm{\theta}}(\mathbf{\hat{x}}_i'') \neq y_i \}.
\end{align}

\vspace{1ex} \noindent Then we can define our adversarial risk separately for non-boundary samples $\mathcal{S}_{f_{\bm{\theta}}}^{\bullet}$, boundary samples $\mathcal{S}_{f_{\bm{\theta}}}^{\circ}$ and misclassified samples $\mathcal{S}_{f_{\bm{\theta}}}^{-}$. Notice that we have $\mathcal{S}_{f_{\bm{\theta}}}^{\bullet} \cup \hspace{0.1em} \mathcal{S}_{f_{\bm{\theta}}}^{\circ} \cup \hspace{0.1em} \mathcal{S}_{f_{\bm{\theta}}}^{-} = \mathcal{S}_{f_{\bm{\theta}}}^{+} \cup \hspace{0.1em} \mathcal{S}_{f_{\bm{\theta}}}^{-} = \mathcal{S}$. First of all, we consider two kinds of risks for non-boundary samples. For one thing, as observed in Figure~\ref{fig:motivation_boundary_samples}, to learn the basic adversarial robustness, it is vital that they are fully perturbed. For another, as we suggest in Section~\ref{subsec:DICAT}, additionally considering a domain interpolation consistency risk helps to learn better generalization. So for $\mathbf{x}_i \in \mathcal{S}_{f_{\bm{\theta}}}^{\bullet}$, given two different augmentations $\mathbf{\dot{x}}_i$ and $\mathbf{\ddot{x}}_i$, we define its adversarial risk as:
\begin{equation}\label{eq:ours_risk_1}
    \mathcal{R}^{\bullet}(f_{\bm{\theta}}, \mathbf{x}_i) := \, \mathds{1}(f_{\bm{\theta}}(\mathbf{\hat{x}}_i') \neq y_i) + \, \mathds{1} \Big( f_{\bm{\theta}} \Big( \frac{\mathbf{\hat{\dot{x}}}_i' + \mathbf{\hat{\ddot{x}}}_i'}{2} \Big) \neq \Big( \frac{f_{\bm{\theta}}(\mathbf{\dot{x}}_i) + f_{\bm{\theta}}(\mathbf{\ddot{x}}_i)}{2}\Big)\Big).
\end{equation}

\vspace{1ex} \noindent In contrast, as we suggest in Section~\ref{subsec:boundary_sample}, applying perturbations of different intensities to boundary samples can hardly impact the final robustness, while moderate perturbations for them do benefit the generalization on clean data. Accordingly, we formulate the adversarial risk for $\mathbf{x}_i \in \mathcal{S}_{f_{\bm{\theta}}}^{\circ}$ as:
\begin{equation}\label{eq:ours_risk_2}
    \mathcal{R}^{\circ}(f_{\bm{\theta}}, \mathbf{x}_i) := \, \mathds{1}(f_{\bm{\theta}}(\mathbf{\hat{x}}_i'') \neq y_i) + \, \mathds{1} \Big( f_{\bm{\theta}} \Big( \frac{\mathbf{\hat{\dot{x}}}_i' + \mathbf{\hat{\ddot{x}}}_i'}{2} \Big) \neq \Big( \frac{f_{\bm{\theta}}(\mathbf{\dot{x}}_i) + f_{\bm{\theta}}(\mathbf{\ddot{x}}_i)}{2}\Big)\Big).
\end{equation}

\vspace{1ex} \noindent Finally, this work does not involve improvement, as well as any specific requirements, for the misclassified samples. So for $\mathbf{x}_i \in \mathcal{S}_{f_{\bm{\theta}}}^{-}$, both directly using the standard adversarial risk as given in Equation~(\ref{eq:adv_risk_standard}) and adopting advanced ones such as $\mathcal{R}^{-}(f_{\bm{\theta}}, \mathbf{x}_i)^{\text{MMA}} = \mathds{1}(f_{\bm{\theta}}(\mathbf{x}_i) \neq y_i)$ suggested by MMA and $\mathcal{R}^{-}(f_{\bm{\theta}}, \mathbf{x}_i)^{\text{MART}} = \mathds{1}(f_{\bm{\theta}}(\mathbf{\hat{x}}_i') \!\neq\! y_i) + \mathds{1}(f_{\bm{\theta}}(\mathbf{x}_i) \!\neq\! f_{\bm{\theta}}(\mathbf{\hat{x}}_i'))$ proposed by MART would be viable options.

\vspace{1ex} \noindent Combining the three risk components above, we can obtain our novel adversarial risk involving the principle of the proposed \textit{Robust Alignment} to be minimized as follows, based on which we end up with our new RAAT method.
\begin{equation}\label{eq:ours_risk}
\mathcal{R}(f_{\bm{\theta}}) := \,\, \sum_{\mathbf{x}_i \in \mathcal{S}_{f_{\bm{\theta}}}^{\bullet}} \! \mathcal{R}^{\bullet}(f_{\bm{\theta}}, \mathbf{x}_i) + \! \sum_{\mathbf{x}_i \in \mathcal{S}_{f_{\bm{\theta}}}^{\circ}} \! \mathcal{R}^{\circ}(f_{\bm{\theta}}, \mathbf{x}_i) + \! \sum_{\mathbf{x}_i \in \mathcal{S}_{f_{\bm{\theta}}}^{-}} \! \mathcal{R}^{-}(f_{\bm{\theta}}, \mathbf{x}_i).
\end{equation}

\subsection{Specific Method Proposed: RAAT \& RAAT$^{++}$}\label{app:proposed_defense}

We now provide our specific methods with surrogate objectives to optimize the new adversarial risk proposed above. Specifically, as the optimization over the \textit{0-1 loss} is conceptual and not computationally tractable, we replace them with proper surrogate loss functions commonly adopted for DNNs to build practical AT algorithms. Firstly, for all the label-based indicator functions (\ie, with input involving label $y_i$, corresponding to a supervised task), a natural surrogate function is \textit{cross-entropy} (CE), which is the common choice for supervised classifications, as well as conventional AT methods. Furthermore, inspired by \citet{carlini2017towards}, MART proposed to use \textit{boosted cross-entropy} (BCE), which appends an additional margin term to CE to improve the decision margin of the classifier. 
Taking $\mathbf{\hat{x}}_i'$ for an example, with $\mathbf{p}_{k} (\mathbf{\hat{x}}_i', \bm{\theta})$ denoting the probability output as provided in Appendix~\ref{subapp:proposed_risk}, these two surrogate losses can be denoted as:
\begin{equation}\label{eq:CE}
    \hspace{-11.5em} \mathcal{L}^{\text{CE}}(\mathbf{p} (\mathbf{\hat{x}}_i', \bm{\theta}), y_i) = -\log \hspace{0.12em} (\mathbf{p}_{y_i} (\mathbf{\hat{x}}_i', \bm{\theta})),
\end{equation}
\begin{equation}\label{eq:BCE}
    \mathcal{L}^{\text{BCE}} (\mathbf{p}(\mathbf{\hat{x}}_i', \bm{\theta}), y_i) = \, \mathcal{L}^{\text{CE}}(\mathbf{p} (\mathbf{\hat{x}}_i', \bm{\theta}), y_i) - \log \hspace{0.15em} (1 - \max_{k \neq y_i} \mathbf{p}_{k} (\mathbf{\hat{x}}_i', \bm{\theta})).
\end{equation}

\vspace{1ex} \noindent Secondly, for the indicator function serving as a consistency regularization term in $\mathcal{R}^{\bullet}(f_{\bm{\theta}}, \mathbf{x}_i)$, we adopt \textit{Jensen-Shannon divergence} (JS) as the surrogate loss. It is a symmetrized and smoothed variant of the commonly used \textit{Kullback–Leibler divergence} (KL), and is suggested by Cons-AT in measuring the difference of output distributions under robust setting. Given $\mathbf{\dot{x}}_i$ and $\mathbf{\ddot{x}}_i$ as in Equation~(\ref{eq:ours_risk_1}) and a parameter $\beta \in (0, 1)$, provided a \textit{mixture distribution} $\mathbf{q} ((\mathbf{\dot{x}}_i, \mathbf{\ddot{x}}_i), \bm{\theta})$ as:
\begin{equation}\label{eq:JS_q}
    \mathbf{q} ((\mathbf{\dot{x}}_i, \mathbf{\ddot{x}}_i), \bm{\theta}) = \, \frac{\hspace{0.1em}1\hspace{0.1em}}{2} \hspace{0.1em} \big( \mathbf{p} ((\beta \cdot \mathbf{\hat{\dot{x}}}_i' + (1 - \beta) \cdot \mathbf{\hat{\ddot{x}}}_i'), \bm{\theta}) + (\beta \cdot \mathbf{p} (\mathbf{\dot{x}}_i, \bm{\theta}) + (1 - \beta) \cdot \mathbf{p} (\mathbf{\ddot{x}}_i, \bm{\theta})) \big),
\end{equation}
and $\mathcal{L}^{\text{KL}}(\cdot \hspace{0.05em} || \hspace{0.05em} \cdot)$ denoting the standard KL loss, then the JS consistency loss adopted in our method is:
\begin{equation}\label{eq:JS}
\begin{aligned}
& \hspace{-0.4em} \mathcal{L}^{\text{JS}}(\mathbf{p} ((\beta \cdot \mathbf{\hat{\dot{x}}}_i' + (1 \!-\! \beta) \cdot \mathbf{\hat{\ddot{x}}}_i'), \bm{\theta}) \hspace{0.1em} || \hspace{0.1em} (\beta \cdot \mathbf{p} (\mathbf{\dot{x}}_i, \bm{\theta}) + (1 \!-\! \beta) \cdot \mathbf{p} (\mathbf{\ddot{x}}_i, \bm{\theta}))) \\
& \hspace{0.7em} = \frac{\hspace{0.05em}1\hspace{0.05em}}{2} \hspace{0.05em} \big( \mathcal{L}^{\text{KL}}( \mathbf{p} ((\beta \cdot \mathbf{\hat{\dot{x}}}_i' + (1 \!-\! \beta) \cdot \mathbf{\hat{\ddot{x}}}_i'), \bm{\theta}) \hspace{0.1em} || \hspace{0.1em} \mathbf{q} ((\mathbf{\dot{x}}_i, \mathbf{\ddot{x}}_i), \bm{\theta})) + \, \mathcal{L}^{\text{KL}}((\beta \cdot \mathbf{p} (\mathbf{\dot{x}}_i, \bm{\theta}) + (1 \!-\! \beta) \cdot \mathbf{p} (\mathbf{\ddot{x}}_i, \bm{\theta})) \hspace{0.1em} || \hspace{0.1em} \mathbf{q} ((\mathbf{\dot{x}}_i, \mathbf{\ddot{x}}_i), \bm{\theta})) \big).
\end{aligned}
\end{equation}

\noindent \textbf{The overall objective.} $\hspace{0.05em}$ Based on these surrogate loss functions, now we can state the overall objective function of our proposed method, \textit{\textbf{R}obust \textbf{A}lignment \textbf{A}dversarial \textbf{T}raining} (\textbf{RAAT}). Note that for the misclassified part of the adversarial risk, according to the particular type of $\mathcal{R}^{-}(f_{\bm{\theta}}, \mathbf{x}_i)$ adopted, we can follow the corresponding original work proposing it to determine the specific surrogate loss to be used. Together with the different types of \textit{cross-entropy}, this leaves room for our surrogate algorithms to be flexibly customized to fit different real-world practices better. Representatively, we provide basic RAAT and advanced RAAT$^{++}$ methods, with their objectives $\mathcal{L}^{\text{RAAT}}(\mathbf{x}_i, y_i, \bm{\theta})$ and $\mathcal{L}^{\text{RAAT}^{++}}(\mathbf{x}_i, y_i, \bm{\theta})$ respectively provided in Table~\ref{tab:losses}, where $\lambda$ is a weight parameter to balance the supervised learning and consistency regularization term, and $\bm{\bar{\theta}}$ is a fixed copy of $\bm{\theta}$, which means the corresponding terms are just used as indicators instead of parts of the optimization objective. As suggested by \citet{zhang2019mixup}, the value of $\beta$ follows \textit{beta distribution} as $\beta \sim Beta(\gamma, \gamma)$ with the parameter $\gamma \in (0, \infty)$. Finally, through minimizing $\mathcal{L}^{\text{RAAT}}(\bm{\theta}) = \frac{1}{n} \sum_{i=1}^{n} \mathcal{L}^{\text{RAAT}}(\mathbf{x}_i, y_i, \bm{\theta})$ or $\mathcal{L}^{\text{RAAT}^{++}}(\bm{\theta}) = \frac{1}{n} \sum_{i=1}^{n} \mathcal{L}^{\text{RAAT}^{++}}(\mathbf{x}_i, y_i, \bm{\theta})$, we are expecting to train a DNN $\bm{\theta}$ with aligned input and latent spaces to achieve better accuracy and robustness at the same time.

\begin{table*}[h!]
\caption{The overall optimization objectives of the proposed RAAT and RAAT$^{++}$, as well as the benchmarks involved in this work. Note that as mentioned in Appendix~\ref{subapp:sm_setup}, different from other methods, the adversarial sample $\mathbf{\hat{x}}'$ in TRADES is generated by maximizing its KL-divergence regularization term. Also, in practice, it is possible for Cons-AT to further utilize the benign augmentations (\ie, $\mathbf{\dot{x}}_i$ and $\mathbf{\ddot{x}}_i$) in the supervised term of the loss, so do our methods.}\label{tab:losses}
\centering
\renewcommand\arraystretch{1.6}
\resizebox{\textwidth}{!}{
    \begin{tabular}{ll}
    \toprule
    \textbf{AT Method} & \textbf{Optimization Objective ($loss$)} \\
    \midrule
    PGD-AT & $\mathcal{L}^{\text{CE}}(\mathbf{p}(\hat{\mathbf{x}}^{\prime},\boldsymbol{\theta}),y)$ \\
    TRADES & $\mathcal{L}^{\text{CE}}(\mathbf{p}(\mathbf{x},\boldsymbol{\theta}),y)+\lambda\cdot\mathcal{L}^{\text{KL}}(\mathbf{p}(\mathbf{x},\boldsymbol{\theta})||\mathbf{p}(\hat{\mathbf{x}}^{\prime},\boldsymbol{\theta}))$ \\
    MART & $\mathcal{L}^{\text{BCE}}(\mathbf{p}(\hat{\mathbf{x}}^{\prime},\boldsymbol{\theta}),y)+\lambda\cdot\mathcal{L}^{\text{KL}}(\mathbf{p}(\mathbf{x},\boldsymbol{\theta})||\mathbf{p}(\hat{\mathbf{x}}^{\prime},\boldsymbol{\theta}))\cdot(1-\mathbf{p}_{y}(\mathbf{x},\boldsymbol{\theta}))$ \\
    Cons-AT & $\mathcal{L}^{\text{CE}}(\mathbf{p}(\hat{\mathbf{x}}^{\prime},\boldsymbol{\theta}),y) + \lambda \cdot \mathcal{L}^{\text{JS}}(\mathbf{p} (\mathbf{\hat{\dot{x}}}_i', \bm{\theta}) \hspace{0.1em} || \hspace{0.1em} \mathbf{p} (\mathbf{\hat{\ddot{x}}}_i', \bm{\theta}))$ \\
    \cmidrule(r){1-2}
    \textbf{RAAT} & $\mathcal{L}^{\text{CE}}(\mathbf{p} (\mathbf{\hat{x}}_i', \bm{\theta}), y_i) \cdot \mathds{1}(h_{\bm{\bar{\theta}}}(\mathbf{\hat{x}}_i'') \!=\! y_i) + \mathcal{L}^{\text{CE}}(\mathbf{p} (\mathbf{\hat{x}}_i'', \bm{\theta}), y_i) \cdot \mathds{1}(h_{\bm{\bar{\theta}}}(\mathbf{\hat{x}}_i'') \!\neq\! y_i)$ \\
    & $+ \lambda \cdot \mathcal{L}^{\text{JS}}(\mathbf{p} ((\beta \!\cdot\! \mathbf{\hat{\dot{x}}}_i' \!+\! (1 \!-\! \beta) \!\cdot\! \mathbf{\hat{\ddot{x}}}_i'), \bm{\theta}) \hspace{0.1em} || \hspace{0.1em} (\beta \!\cdot\! \mathbf{p} (\mathbf{\dot{x}}_i, \bm{\theta}) \!+\! (1 \!-\! \beta) \!\cdot\! \mathbf{p} (\mathbf{\ddot{x}}_i, \bm{\theta}))) \cdot \mathds{1}(h_{\bm{\bar{\theta}}}(\mathbf{x}_i) \!=\! y_i)$ \\
    \cmidrule(r){2-2}
    \textbf{RAAT$^{++}$} & $\mathcal{L}^{\text{BCE}}(\mathbf{p} (\mathbf{\hat{x}}_i', \bm{\theta}), y_i) \cdot \mathds{1}(h_{\bm{\bar{\theta}}}(\mathbf{\hat{x}}_i'') \!=\! y_i) + \mathcal{L}^{\text{BCE}}(\mathbf{p} (\mathbf{\hat{x}}_i'', \bm{\theta}), y_i) \cdot \mathds{1}(h_{\bm{\bar{\theta}}}(\mathbf{\hat{x}}_i'') \!\neq\! y_i)$ \\
    & $+ \lambda \cdot \big( \mathcal{L}^{\text{JS}}(\mathbf{p} ((\beta \!\cdot\! \mathbf{\hat{\dot{x}}}_i' \!+\! (1 \!-\! \beta) \!\cdot\! \mathbf{\hat{\ddot{x}}}_i'), \bm{\theta}) \hspace{0.1em} || \hspace{0.1em} (\beta \!\cdot\! \mathbf{p} (\mathbf{\dot{x}}_i, \bm{\theta}) \!+\! (1 \!-\! \beta) \!\cdot\! \mathbf{p} (\mathbf{\ddot{x}}_i, \bm{\theta}))) \cdot \mathds{1}(h_{\bm{\bar{\theta}}}(\mathbf{x}_i) \!=\! y_i) + \mathcal{L}^{\text{KL}}(\mathbf{p}(\mathbf{x}_i, \bm{\theta}) \hspace{0.1em} || \hspace{0.1em} \mathbf{p}(\mathbf{\hat{x}}_i', \bm{\theta})) \cdot (1 \!-\! \mathbf{p}_{y_i} (\mathbf{x}_i, \bm{\theta})) \big)$ \\
    \bottomrule
    \end{tabular}
}
\end{table*}

\section{Additional Experimental Details}\label{app:exp}

\subsection{Experimental Setup}\label{subapp:sm_setup}

Following \citet{pang2021bag}, which explores a great number of hyper-parameter settings for AT, we adopt the following common settings. For outer minimization, we use SGD optimizer with momentum 0.9, batch size 128, weight decay $5 \!\times\! 10^{-4}$, initial learning rate 0.1, total 110 training epochs with learning rate decay by a factor of 0.1 at 100 and 105 epochs, respectively. The only exception is in Section~\ref{subsec:estimate}, where we adopt 200 training epochs with learning rate decay at 100 and 150 epochs to strictly ensure fairness in comparison with the current SOTAs. For the inner maximization, under the $\ell_{\infty}$ threat model with maximal perturbation budget $\epsilon = 8/255$, we adopt step size $\alpha = 2/255$ with 10-step adversaries (\ie, PGD-10 except for TRADES, in which the adversarial samples are generated by maximizing its KL-divergence regularization term~\citep{zhang2019theoretically}). While under the $\ell_2$ threat model with maximal perturbation budget $\epsilon = 128/255$, we adopt step size $\alpha = 32/255$. Same as their original papers, we set the regularization parameter $\lambda = 6$ for TRADES and MART, as well as $\lambda = 1$ for Cons-AT and ours. Also, we use default $\eta = 0.1$ to acquire boundary samples and fix $\gamma = 0.75$ for the \textit{beta distribution}. For data pre-processing, we normalize all natural images into [0, 1], and adopt standard data augmentations including random crop with 4-pixel zero padding and random horizontal flip with 50\% of probability.
% Also, the temperature in Cons-AT and our method is fixed to $\tau = 0.5$.

\vspace{1ex} \noindent The experiments are conducted on Ubuntu 22.04 OS with Intel Xeon Platinum 8336C 32-Core 2.3GHz CPU, 512GB RAM and 8 $\!\times\!$ NVIDIA GeForce RTX 4090 GPUs, and are implemented with Python 3.8.19 and PyTorch 1.8.1 $\!$+$\!$ cu111.

\subsection{Ablation Study}\label{subapp:ablation}

\begin{figure*}[h!]
  \centering
  \subfloat[\small Ablation of two ideas.]
  {\includegraphics[width=0.333\textwidth]{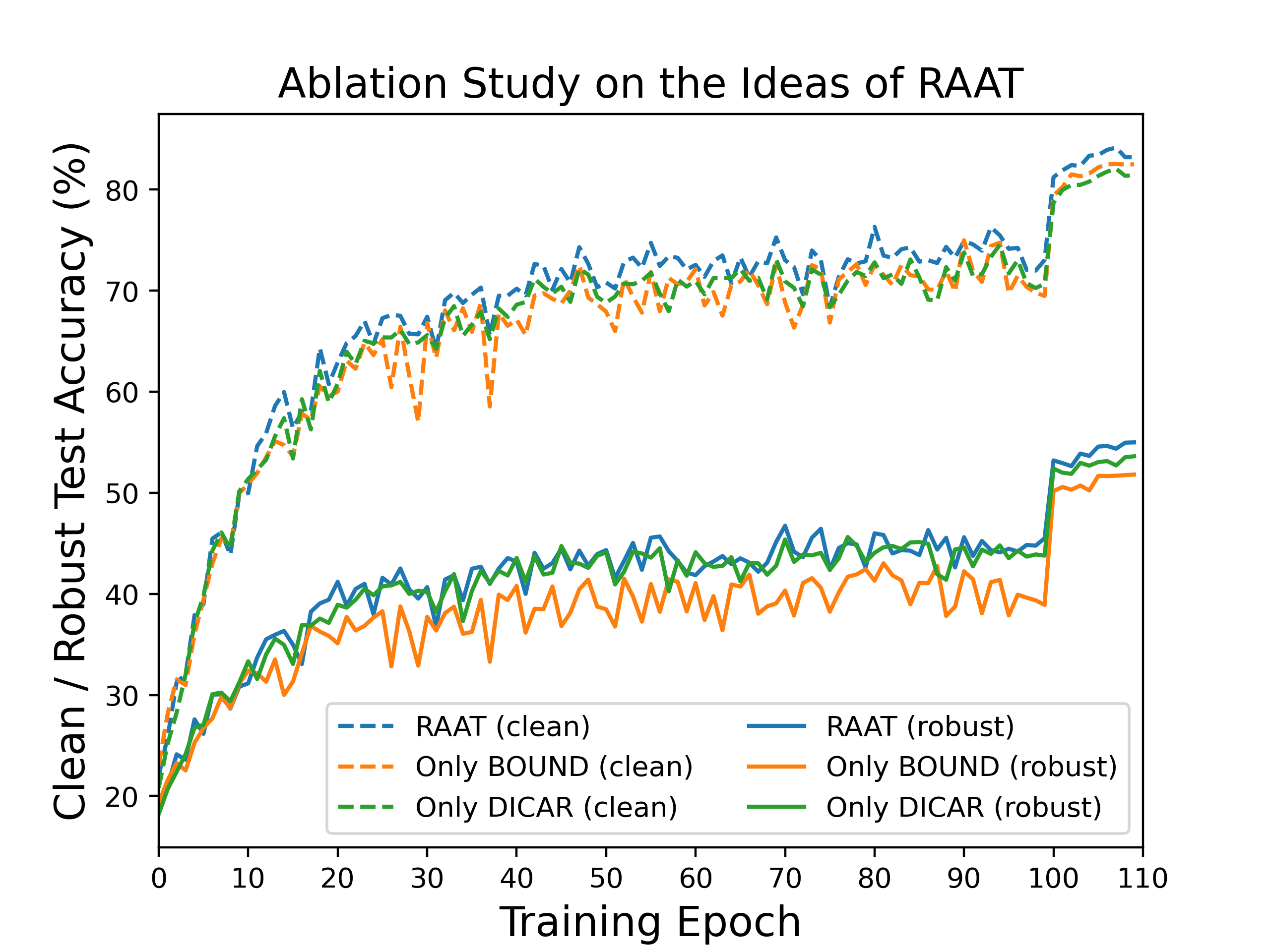}}
  \hspace{-0.9em}
  \subfloat[\small Impact of boundary range.]
  {\includegraphics[width=0.333\textwidth]{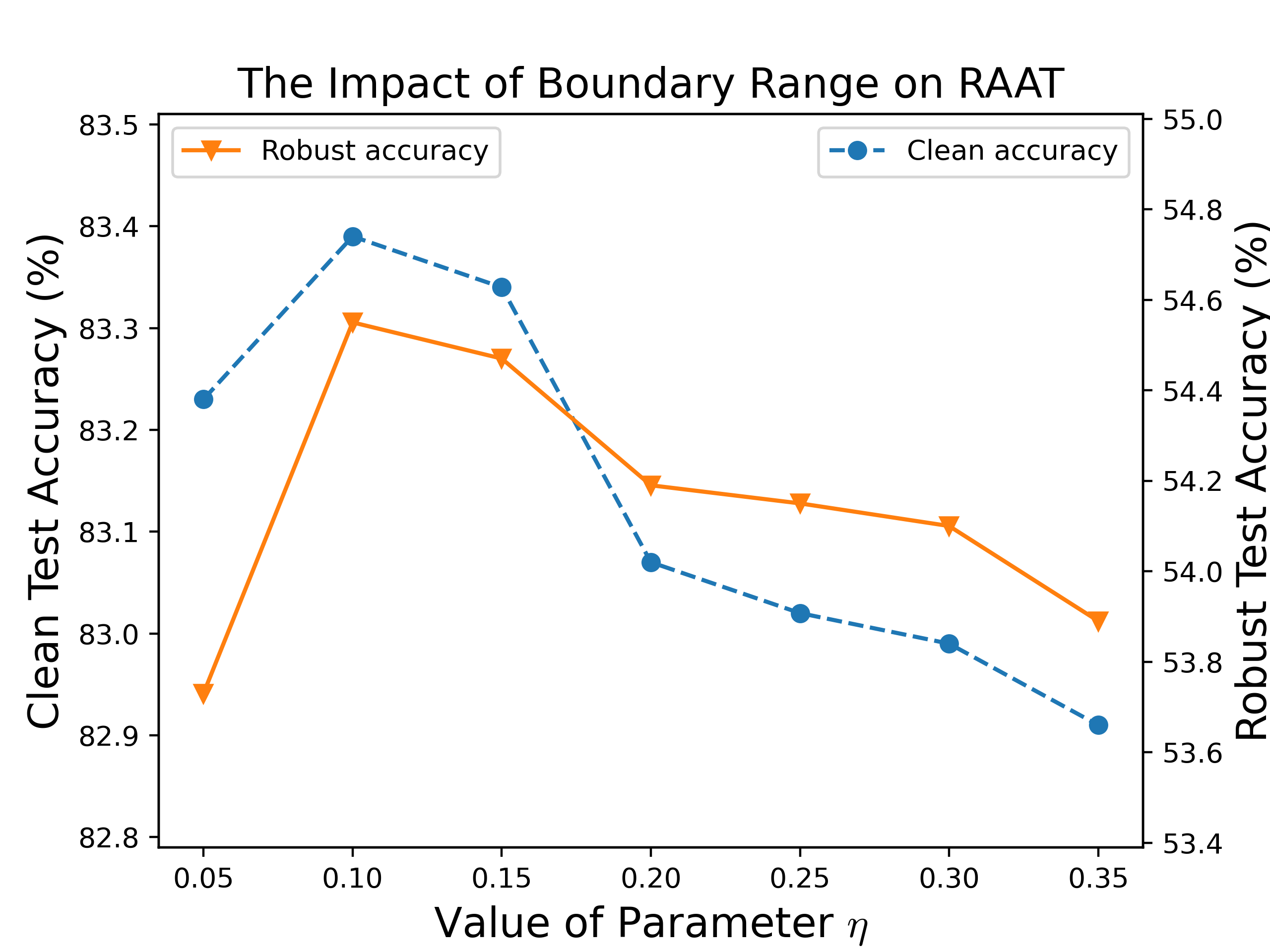}}
  \hspace{0.1em}
  \subfloat[\small Sensitivity to regularization weight.]
  {\includegraphics[width=0.333\textwidth]{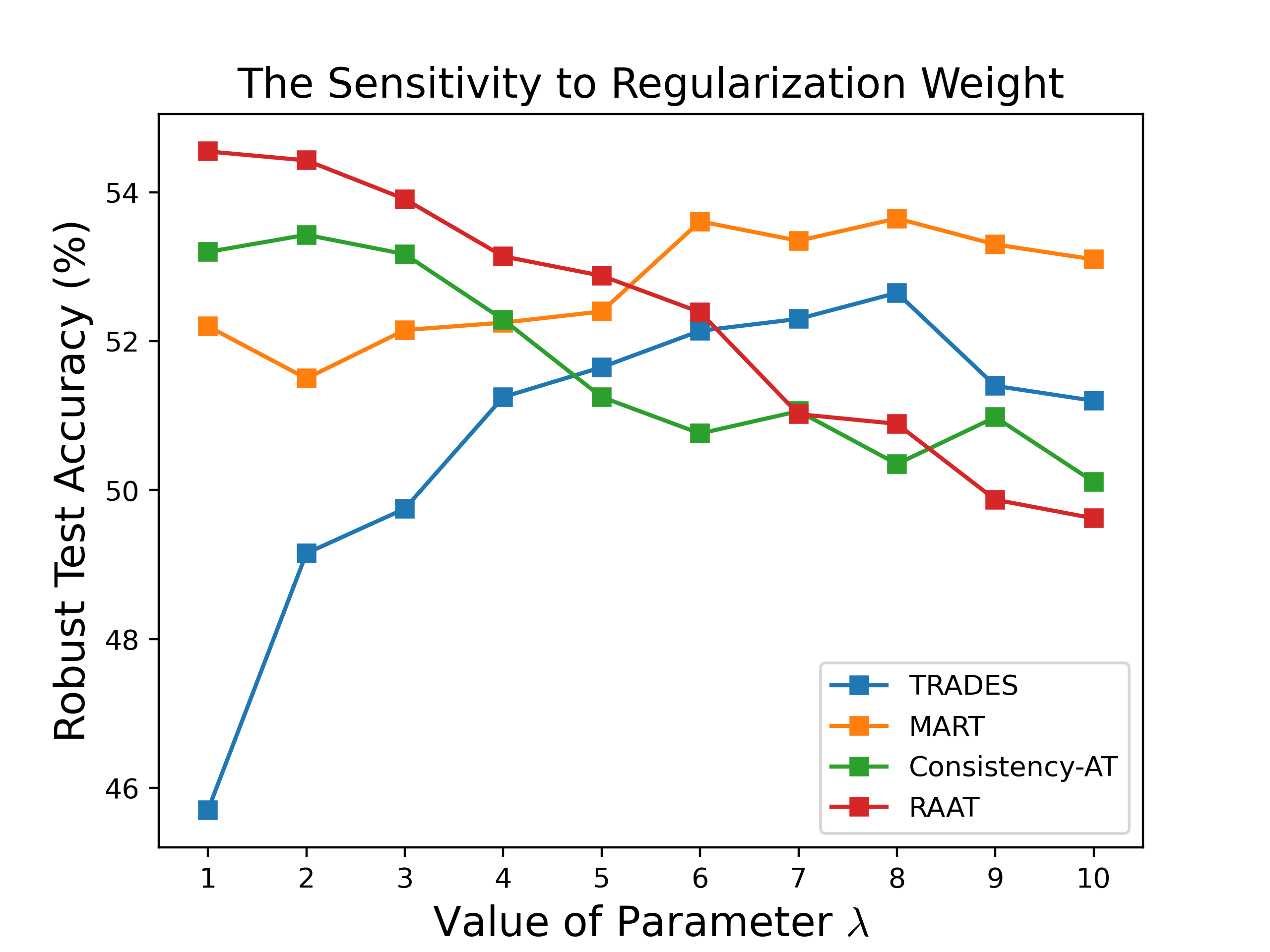}}
  \caption{Three comprehensive ablation experiments of the proposed RAAT on CIFAR-10. The blue line in (a) and the points respectively indicating $\eta = 0.1$ and $\lambda = 1$ in (b) and (c) correspond to the RAAT record under ResNet-18 in Table~\ref{tab:combined}.}
  \label{fig:ablation}
\end{figure*}

To better examine the effectiveness of the proposed defense, we investigate RAAT from the following perspectives: \uppercase\expandafter{\romannumeral1}) the effectiveness of the two ideas adopted by RAAT (Figure~\ref{fig:ablation} (a)); \uppercase\expandafter{\romannumeral2}) the impact of the specific boundary range, which is determined by $\eta$ (Figure~\ref{fig:ablation} (b)); and \uppercase\expandafter{\romannumeral3}) the sensitivity to the regularization parameter $\lambda$ (Figure~\ref{fig:ablation} (c)).

\begin{table*}[h]
\caption{Experimental results on CIFAR-10, CIFAR-100, and Tiny-ImageNet datasets with ResNet-18 architecture under the $\ell_{2}$ threat model. All the results are averages of three runs. The final row for each dataset marked with $\uparrow$ (\%) shows the percentage improvement of the best of RAAT/RAAT$^{++}$ over the best of the four benchmarks.}\label{tab:l2}
  \centering
  \renewcommand\arraystretch{0.95}
  \definecolor{mygreen}{rgb}{0,0.6,0}
  \definecolor{Salmon}{rgb}{0.98,0.5,0.45}
  \definecolor{Gray}{rgb}{0.5,0.5,0.5}
  \resizebox{0.65\textwidth}{!}{%
  \begin{tabular}{clccccc}
    \toprule
    \multirow{2.5}{*}{\textbf{Dataset}} & \multirow{2.5}{*}{\textbf{Method}} & \multicolumn{5}{c}{\textbf{ResNet-18}} \\
    \cmidrule(lr){3-7}
    & & \textbf{Clean} & \textbf{PGD-10} & \textbf{PGD-100} & \textbf{C\&W} & \textbf{AA} \\
    \midrule
    \multirow{7}{*}{\rotatebox{90}{\textbf{CIFAR-10}}} 
    & PGD-AT & 87.76\tiny{$\pm$0.51} & 68.82\tiny{$\pm$0.23} & 67.72\tiny{$\pm$0.23} & 66.64\tiny{$\pm$0.43} & 66.36\tiny{$\pm$0.37} \\
    & TRADES & 83.99\tiny{$\pm$0.28} & 68.74\tiny{$\pm$0.32} & 68.59\tiny{$\pm$0.63} & 67.05\tiny{$\pm$0.55} & 65.93\tiny{$\pm$0.30} \\
    & MART & 84.09\tiny{$\pm$0.19} & 68.96\tiny{$\pm$0.37} & 68.19\tiny{$\pm$0.21} & 67.22\tiny{$\pm$0.43} & 66.28\tiny{$\pm$0.27} \\
    & Cons-AT & 88.76\tiny{$\pm$0.41} & 70.21\tiny{$\pm$0.49} & 69.16\tiny{$\pm$0.45} & 68.10\tiny{$\pm$0.42} & 67.46\tiny{$\pm$0.26} \\
    & \cellcolor{black!10}\textbf{RAAT} & \cellcolor{black!10}\textbf{89.17}\tiny{$\pm$0.26} & \cellcolor{black!10}69.76\tiny{$\pm$0.35} & \cellcolor{black!10}68.67\tiny{$\pm$0.17} & \cellcolor{black!10}\textbf{68.13}\tiny{$\pm$0.30} & \cellcolor{black!10}67.95\tiny{$\pm$0.24} \\
    & \cellcolor{black!10}\textbf{RAAT$^{++}$} & \cellcolor{black!10}86.54\tiny{$\pm$0.33} & \cellcolor{black!10}\textbf{70.84}\tiny{$\pm$0.17} & \cellcolor{black!10}\textbf{70.01}\tiny{$\pm$0.15} & \cellcolor{black!10}69.07\tiny{$\pm$0.28} & \cellcolor{black!10}\textbf{67.96}\tiny{$\pm$0.24} \\
    \cmidrule(lr){2-7}
    & \hspace{0.5em}\textbf{$\uparrow$ (\%)} & \textcolor{mygreen}{+0.46\%} & \textcolor{mygreen}{+0.90\%} & \textcolor{mygreen}{+1.23\%} & \textcolor{mygreen}{+0.04\%} & \textcolor{mygreen}{+0.74\%} \\
    \midrule
    \multirow{7}{*}{\rotatebox{90}{\textbf{CIFAR-100}}} 
    & PGD-AT & 65.00\tiny{$\pm$0.41} & 42.47\tiny{$\pm$0.29} & 41.29\tiny{$\pm$0.18} & 40.03\tiny{$\pm$0.17} & 39.27\tiny{$\pm$0.39} \\
    & TRADES & 61.25\tiny{$\pm$0.48} & 43.52\tiny{$\pm$0.58} & 42.96\tiny{$\pm$0.19} & 41.22\tiny{$\pm$0.38} & 40.15\tiny{$\pm$0.42} \\
    & MART & 60.08\tiny{$\pm$0.19} & 43.92\tiny{$\pm$0.16} & 43.28\tiny{$\pm$0.35} & 41.63\tiny{$\pm$0.33} & 39.85\tiny{$\pm$0.20} \\
    & Cons-AT & 65.14\tiny{$\pm$0.30} & 43.16\tiny{$\pm$0.37} & 42.09\tiny{$\pm$0.32} & 40.99\tiny{$\pm$0.36} & 39.92\tiny{$\pm$0.26} \\
    & \cellcolor{black!10}\textbf{RAAT} & \cellcolor{black!10}\textbf{66.12}\tiny{$\pm$0.11} & \cellcolor{black!10}43.47\tiny{$\pm$0.20} & \cellcolor{black!10}42.22\tiny{$\pm$0.26} & \cellcolor{black!10}41.33\tiny{$\pm$0.17} & \cellcolor{black!10}40.25\tiny{$\pm$0.20} \\
    & \cellcolor{black!10}\textbf{RAAT$^{++}$} & \cellcolor{black!10}62.43\tiny{$\pm$0.09} & \cellcolor{black!10}\textbf{44.95}\tiny{$\pm$0.15} & \cellcolor{black!10}\textbf{43.96}\tiny{$\pm$0.12} & \cellcolor{black!10}\textbf{42.35}\tiny{$\pm$0.23} & \cellcolor{black!10}\textbf{40.61}\tiny{$\pm$0.15} \\
    \cmidrule(lr){2-7}
    & \hspace{0.5em}\textbf{$\uparrow$ (\%)} & \textcolor{mygreen}{+1.50\%} & \textcolor{mygreen}{+2.34\%} & \textcolor{mygreen}{+1.57\%} & \textcolor{mygreen}{+1.73\%} & \textcolor{mygreen}{+0.90\%} \\
    \midrule
    \multirow{7}{*}{\rotatebox{90}{\textbf{Tiny-ImageNet}}} 
    & PGD-AT & 59.36\tiny{$\pm$0.18} & 43.52\tiny{$\pm$0.16} & 43.05\tiny{$\pm$0.20} & 42.81\tiny{$\pm$0.18} & 40.90\tiny{$\pm$0.24} \\
    & TRADES & 57.61\tiny{$\pm$0.44} & 44.84\tiny{$\pm$0.42} & 44.65\tiny{$\pm$0.54} & 43.23\tiny{$\pm$0.21} & 42.33\tiny{$\pm$0.27} \\
    & MART & 56.47\tiny{$\pm$0.22} & 45.05\tiny{$\pm$0.23} & 44.77\tiny{$\pm$0.16} & 43.10\tiny{$\pm$0.15} & 42.22\tiny{$\pm$0.16} \\
    & Cons-AT & \textbf{61.49}\tiny{$\pm$0.24} & 45.32\tiny{$\pm$0.28} & 44.85\tiny{$\pm$0.19} & 43.57\tiny{$\pm$0.30} & 42.59\tiny{$\pm$0.25} \\
    & \cellcolor{black!10}\textbf{RAAT} & \cellcolor{black!10}61.20\tiny{$\pm$0.12} & \cellcolor{black!10}\textbf{45.43}\tiny{$\pm$0.13} & \cellcolor{black!10}44.66\tiny{$\pm$0.19} & \cellcolor{black!10}\textbf{43.82}\tiny{$\pm$0.13} & \cellcolor{black!10}\textbf{42.63}\tiny{$\pm$0.11} \\
    & \cellcolor{black!10}\textbf{RAAT$^{++}$} & \cellcolor{black!10}57.95\tiny{$\pm$0.10} & \cellcolor{black!10}45.38\tiny{$\pm$0.12} & \cellcolor{black!10}\textbf{44.92}\tiny{$\pm$0.12} & \cellcolor{black!10}43.61\tiny{$\pm$0.08} & \cellcolor{black!10}42.35\tiny{$\pm$0.11} \\
    \cmidrule(lr){2-7}
    & \hspace{0.5em}\textbf{$\uparrow$ (\%)} & \textcolor{Salmon}{-0.47\%} & \textcolor{mygreen}{+0.24\%} & \textcolor{mygreen}{+0.16\%} & \textcolor{mygreen}{+0.57\%} & \textcolor{mygreen}{+0.09\%} \\
    \bottomrule
  \end{tabular}
  }
\end{table*}

\vspace{1ex} \noindent Firstly, as we expected, both reducing perturbation for boundary samples (referred to as ``BOUND'') and adding the DICAR term benefit clean accuracy and robustness at the same time, concurrently contributing to the final effectiveness of RAAT in improving the current trade-off problem. Secondly, the experiments on $\eta$ confirm our findings in Figure~\ref{fig:motivation_boundary_samples} (c) and (f) that an appropriate partition of boundary and non-boundary samples helps achieve better performance on both perturbed and clean data. Although $\eta$ is fixed at 0.1 in our experiments for simplicity, we have no objection to fine-tuning in real-world practices. It might contain certain underlying significance and bring extra improvement especially for datasets with varying class numbers. 

\vspace{1ex} \noindent Finally, we explore various values of the weight $\lambda$ for all the experimental methods having a regularization term. Here we focus on robustness as it is what the adversarial regularization terms mainly impact. Our results show the best $\lambda$ for TRADES and MART is around 6 to 8, aligning with their original papers. In contrast, $\lambda \in [1,3]$ suits Cons-AT and RAAT more. This is probably because, as mentioned at the end of Section~\ref{subsec:DICAT}, they inject stronger consistency-based inductive biases into models than individual adversarial samples-based conventional AT, making them more sensitive to inappropriately large $\lambda$ values. Overall, compared to optimal cases of all these methods among the entire experimental value range, RAAT achieves the best robustness.

\end{document}